\documentclass{selfevolagent}

\usepackage{pifont}
\usepackage{tabularx}
\usepackage{colortbl}
\usepackage{subcaption}
\usepackage{float}
\usepackage[ruled,linesnumbered]{algorithm2e}

\usepackage{amssymb,amsthm}
\newcommand{\mcV}{\mathcal{V}}
\newcommand{\Vsize}{V}
\newcommand{\Cat}{\mathrm{Cat}}
\newcommand{\KL}{\mathrm{KL}}
\newcommand{\E}{\mathbb{E}}
\newcommand{\R}{\mathbb{R}}
\newcommand{\1}{\mathbf{1}}
\newcommand{\I}{\mathbb{I}}
\newcommand{\vx}[1]{\bm{#1}}
\newcommand{\xzero}{\vx{x}_0}
\newcommand{\xt}{\vx{x}_t}

\newcommand{\xprev}{\vx{x}_{t-\Delta t}}
\newcommand{\xth}{\vx{x}_{\theta}}

\newcommand{\Loss}{\mathcal{L}}
\newcommand{\alphat}{\alpha_t}
\newcommand{\alphap}{\alpha_t'}
\definecolor{DarkGreen}{rgb}{0.0, 0.4, 0}
\definecolor{ludigreen}{RGB}{232,248,232}
\providecommand{\best}[1]{\textbf{#1}}
\providecommand{\second}[1]{\underline{#1}}

\title{Less Uniform Discrete Diffusion is More Powerful and Scalable}
\author[1,2]{Kaibo Wang\cofirst}
\author[1,2]{Ding Ding\cofirst}
\author[1,2]{Fangyu Ding}
\author[2]{Zijin Feng}
\author[2]{Han Shi}
\author[2]{Haoli Bai}
\author[2]{Jiacheng Sun\corrauthor}
\author[1]{Yang Xiang\corrauthor}
\affiliation[1]{The Hong Kong University of Science and Technology}
\affiliation[2]{Huawei Foundation Model Department}
\code{\url{https://github.com/FMD-NEXT/LUDI}}

\abstract{Although uniform diffusion language models (UDLMs) represent a promising diffusion paradigm, scaling them remains challenging. We identify the core obstacle as an over-uniform training objective and condition-target confusion during sampling. To address these, we propose Less Uniform Diffusion (LUDI), a novel UDLM framework. Specifically, we (i) introduce a less uniform loss that directs each reverse transition toward the clean token, and (ii) equip the model with per-token time embeddings that supply token-level corruption hints, enabling confidence-based few-step sampling. Experiments across scales show that LUDI yields cleaner supervision and improves few-step generation. We further continue-train a 7B autoregressive model into LUDI-7B, resulting in a UDLM capable of complex reasoning. It achieves a 3-token-per-step speedup over AR decoding and competitive performance compared with masked diffusion baselines, revealing that the full potential of UDLMs for complex generation remains to be unlocked.}

\begin{document}
\maketitle
\begingroup
\renewcommand{\thefootnote}{}
\footnotetext{\noindent
\begin{tabular}{@{}r@{\hspace{0.45em}}l@{}}
\textsuperscript{\textdagger} & Equal contribution: \email{kwangbi@connect.ust.hk} and \email{ddingab@connect.ust.hk}.\\[2pt]
\textsuperscript{*} & Corresponding authors: \email{sunjiacheng1@huawei.com} and \email{maxiang@ust.hk}.
\end{tabular}}
\endgroup

\section{Introduction}
\begin{figure}[b]
    \centering
    \begin{subfigure}[t]{0.49\textwidth}
        \centering
        \includegraphics[width=\linewidth]{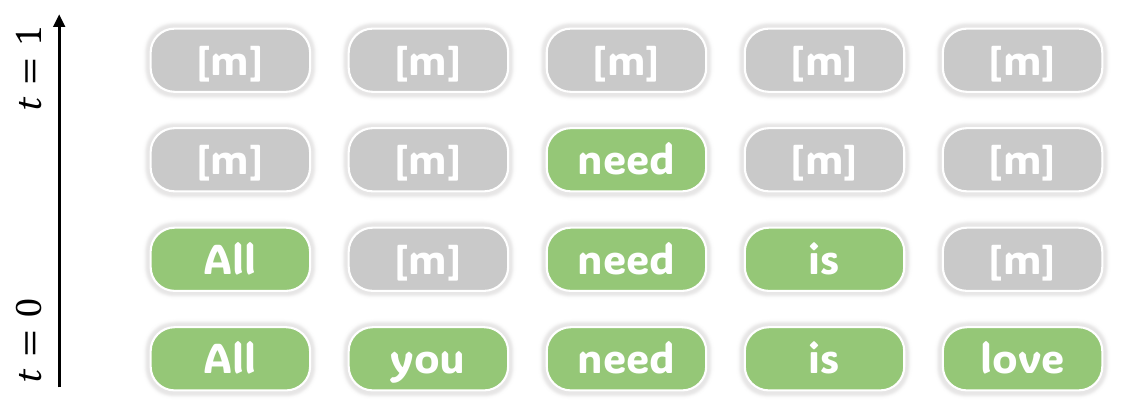}
        \caption{Masked diffusion (MDLM).}
        \label{fig:diffusion-mdlm}
    \end{subfigure}\hfill
    \begin{subfigure}[t]{0.49\textwidth}
        \centering
        \includegraphics[width=\linewidth]{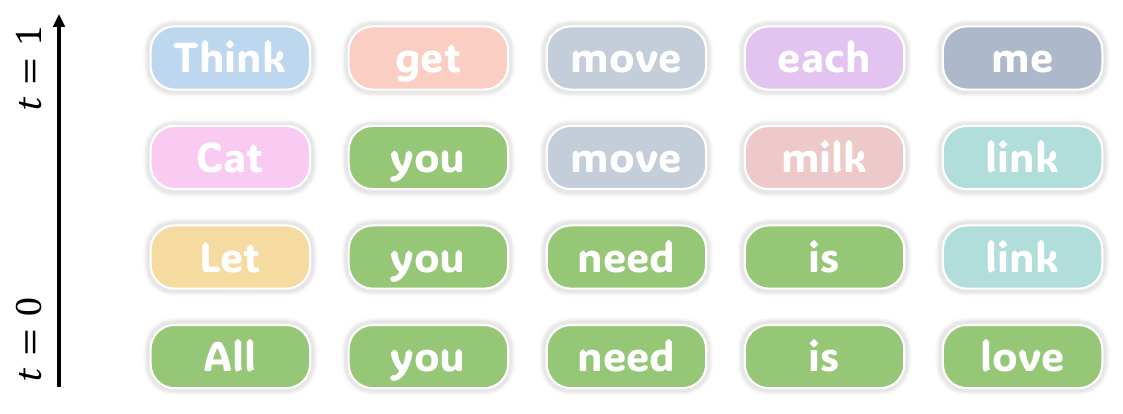}
        \caption{Uniform diffusion (UDLM).}
        \label{fig:diffusion-udlm}
    \end{subfigure}
    \caption{Schematic comparison between masked and uniform diffusion.}
    \label{fig:diffusion-comparison}
\end{figure}

Diffusion large language models (dLLMs) have emerged as competitive alternatives to autoregressive (AR) models, driven by their potential for parallel decoding and bidirectional context modeling~\citep{sahoo2024mdlm, ou2025radd}. Among them, masked diffusion language models (MDLMs, Figure~\ref{fig:diffusion-comparison}\subref{fig:diffusion-mdlm}), which use a mask token as the noise prior, dominate current research, with models at 8B and even 100B scales delivering strong results on tasks like mathematics and programming~\citep{nie2025llada, bie2025llada2}. Uniform diffusion language models (UDLMs, Figure~\ref{fig:diffusion-comparison}\subref{fig:diffusion-udlm}), which instead adopt a uniform prior over the vocabulary, represent another branch that has shown greater potential for few-step generation and self-correction~\citep{sahoo2025duo,schiff2025guidance}. Yet their development lags significantly behind. To date, the advantages of UDLMs have only been demonstrated at small scales, and their viability on complex reasoning tasks remains unestablished.

\begin{figure}[t]
    \centering
    \begin{subfigure}[t]{0.495\textwidth}
        \centering
        \includegraphics[page=1,width=\linewidth]{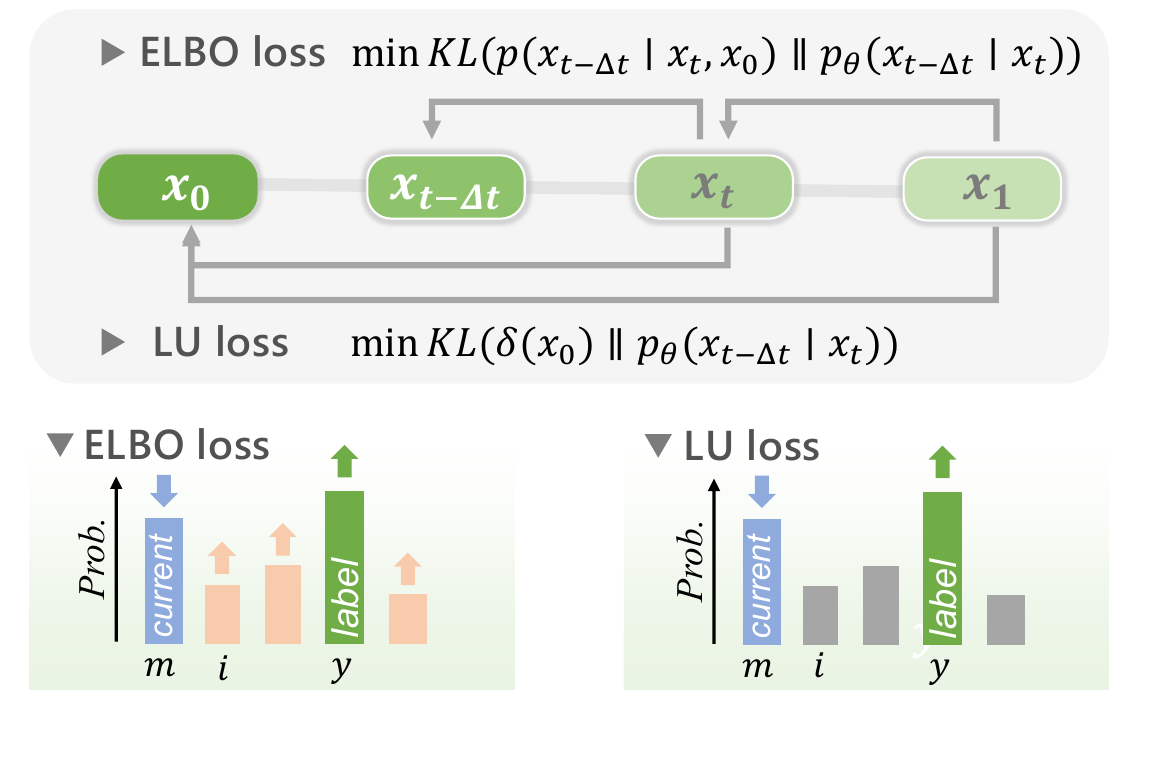}
        \caption{Less uniform loss.}
        \label{fig:ludi-loss}
    \end{subfigure}\hfill
    \begin{subfigure}[t]{0.495\textwidth}
        \centering
        \includegraphics[page=2,width=\linewidth]{figures/ludi_overview_panels.pdf}
        \caption{LUDI with per-token time embeddings.}
        \label{fig:ludi-token-time}
    \end{subfigure}
    \caption{Illustration of Less Uniform Diffusion (LUDI). (a) \textbf{LU Loss.} The ELBO loss (\textit{left}) matches the reverse transition to $p(\mathbf{x}_{t-\Delta t}\mid \mathbf{x}_t,\mathbf{x}_0)$, leading to over-uniform predictions by encouraging all token probabilities ($\sum_i$, \rotatebox{90}{\small\textcolor[HTML]{F8CBAD}{\ding{220}}}). Our LU loss (\textit{right}) removes the smoothing term and retains only the suppression of the erroneous token ($m$ \rotatebox{90}{\small\textcolor[HTML]{8FAADC}{\rotatebox[origin=c]{180}{\ding{220}}}}) and promotion of the target token ($y$ \rotatebox{90}{\small\textcolor[HTML]{70AD47}{\ding{220}}}). (b) \textbf{Per-token time embedding.} The model receives token-level corruption hints via per-token time embeddings. Combined with the LU loss, this allows scaling LUDI to 7B parameters from AR initialization, enabling confidence-based decoding for complex reasoning tasks.}
    \label{fig:ludi-overview}
\end{figure}

Prior efforts mainly focused on conceptually validating the promise of UDLMs. UDLMs can be distilled through connections to continuous Gaussian diffusion~\citep{sahoo2025duo}, trained stably with a simplified loss~\citep{zhu2025sddlm}, and exhibit more favorable scaling behavior under data-constrained conditions~\citep{vonrutte2025scaling}. Nevertheless, existing efforts remain confined to small datasets or scaling-law investigations, leaving a pressing question: \textit{Can UDLMs scale to a larger regime and handle complex reasoning tasks?}

We find that UDLMs do not scale up as naturally as MDLMs. The central difficulty arises from an \textit{over-uniform training objective} and \textit{condition-target confusion} during sampling. During training, the KL-based objective provides an excessively label-smoothed learning signal: the model's predictions become more uniform and less confident in the clean token. During sampling, for complex tasks, simpler tokens should be generated first to serve as conditions, after which the remaining tokens are denoised. In UDLMs, however, all tokens are denoised at a uniform rate. While this works for simple tasks, it leads to condition-target confusion on complex tasks, eventually causing the model to collapse to repeated tokens. These two issues jointly limit the performance of UDLMs on complex tasks: over-uniformity reduces training efficiency and weakens few-step generation capability, while uniformly updating all tokens across many steps leads to condition-target confusion.

To address these issues and make UDLMs practically scalable, we propose \textbf{L}ess \textbf{U}niform \textbf{Di}ffusion (LUDI). LUDI introduces two key modifications. First, by simplifying and approximating the original UDLM training loss, we decompose it into three terms and remove the term that encourages uniform predictions, yielding a less uniform (LU) loss (Figure~\ref{fig:ludi-overview}\subref{fig:ludi-loss}). We prove theoretically that LU loss steers each reverse transition toward the clean token, supporting few-step sampling. Second, to mitigate the confusion between conditions and targets, we provide each token with an individual time embedding during training, which indicates the probability that the token has been randomly replaced (Figure~\ref{fig:ludi-overview}\subref{fig:ludi-token-time}). This allows the model to better distinguish condition tokens from target tokens and naturally enables confidence-based sampling strategies akin to those in MDLMs, facilitating an easy-to-hard generation paradigm.

We conduct extensive experiments to validate LUDI. Pretraining at 170M and 1B scales demonstrates that the LU loss offers cleaner supervision and improves both few-step generation and downstream task performance. We further obtain LUDI-7B by continuing training from an autoregressive checkpoint, a UDLM capable of complex reasoning tasks. Compared with the AR model, LUDI-7B achieves approximately 3 tokens per step while maintaining comparable performance. Compared with MDLMs, LUDI-7B also achieves competitive performance, revealing that the potential of UDLMs for complex generation remains largely untapped. Our contributions are threefold:
\begin{enumerate}
    \item We attribute the difficulty of scaling UDLMs to over-uniform training objectives and condition-target confusion during sampling.
    \item We propose the LUDI loss and per-token time embeddings, enabling UDLMs to perform confidence-based few-step generation.
    \item Our experiments with LUDI-7B establish that UDLMs scale effectively for complex reasoning, delivering a competitive balance between generation quality and speed.
\end{enumerate}

\section{Background}

In this section, we review UDLM. A clean token is corrupted by the forward process into the uniform distribution. By parameterizing and learning the reverse process, samples can be generated by denoising from the uniform prior.

\subsection{Forward Process}

Continuous-time discrete diffusion models corrupt the data distribution $p_0(\mathbf{x})$ into a noise prior through a forward Markov process. In masked diffusion models, the terminal prior is a point mass on the mask token. In uniform-state diffusion models, as $t$ evolves from $0$ to $1$, a clean token $\mathbf{x}_0$ is gradually perturbed toward the uniform distribution over the vocabulary $\mathcal{V}$, where $|\mathcal{V}|=V$. Over an infinitesimal interval from $t$ to $t+\Delta t$, the transition probability of $\mathbf{x}_t$ (ignoring $o(\Delta t)$ terms) is
\begin{equation}
    p(\mathbf{x}_{t+\Delta t}\mid \mathbf{x}_t) = \delta(\mathbf{x}_t,\mathbf{x}_{t+\Delta t}) + Q_t(\mathbf{x}_t,\mathbf{x}_{t+\Delta t})\Delta t .
\end{equation}
Here, $\mathbf{x}_0$ and $\mathbf{x}_t$ are one-hot vectors over $\mathcal{V}$, $\odot$ denotes the Hadamard product, $\delta(\mathbf{x}_t,\mathbf{y})=\mathbb{I}_{\mathbf{y}=\mathbf{x}_t}$, and $Q_t\in\mathbb{R}^{V\times V}$ denotes the transition rate matrix. For uniform-state diffusion and $t\in[0,1)$, the transition rate matrix is given by $Q_t(\mathbf{e}_i,\mathbf{e}_j)
    =
    -\frac{\alpha_t'}{\alpha_t}
    \left(\frac{1}{V}-\mathbb{I}_{i=j}\right)$, where the noise schedule $t\mapsto\alpha_t$ is continuously differentiable and non-increasing, with $\alpha_0=1$ and $\alpha_1=0$. A common choice is $\alpha_t=1-t$. This forward process yields the marginal distribution
\begin{equation}
    p(\mathbf{x}_t\mid \mathbf{x}_0) = \mathrm{Cat}\left( \cdot;\, \alpha_t\mathbf{x}_0+(1-\alpha_t)\frac{\mathbf{1}}{V} \right).
\end{equation}

Let $\alpha_{t\mid s}:=\alpha_t/\alpha_s$ and define $\bar{\mathbf{x}}:=V\alpha_t\mathbf{x}+(1-\alpha_t)\mathbf{1}$. The exact reverse posterior $p(\mathbf{x}_{s}\mid \mathbf{x}_t,\mathbf{x}_0)=\mathrm{Cat}\left(\cdot;\,\boldsymbol{\pi}_{t\to s}\right)$ can be written in closed form:
\begin{equation}
    \label{eq:reverse-posterior}
    \fitbox{0.98\textwidth}{$\displaystyle \boldsymbol{\pi}_{t\to s} = \frac{1}{\langle \bar{\mathbf{x}}_t,\mathbf{x}_0\rangle}
    \Big[V\alpha_t\,\mathbf{x}_t\odot\mathbf{x}_0 +(\alpha_{t\mid s}-\alpha_t)\mathbf{x}_t +(\alpha_{s}-\alpha_t)\mathbf{x}_0 +(1-\alpha_{t\mid
    s})(1-\alpha_{s})\frac{\mathbf{1}}{V} \Big].$}
\end{equation}

We provide a more detailed derivation in Appendix~\ref{app:backward-posteriors}.
\subsection{Reverse Process and Training Objective}

To sample from $p_0(\mathbf{x})$, one learns a reverse denoising process from the noise prior. This requires parameterizing the reverse transition $p_{\theta}(\mathbf{x}_{t-\Delta t}\mid \mathbf{x}_t)$ and matching it to the exact posterior $p(\mathbf{x}_{t-\Delta t}\mid \mathbf{x}_t,\mathbf{x}_0)$. The standard diffusion objective minimizes the local KL divergence
\begin{equation}
    \mathrm{KL}\left( p(\mathbf{x}_{t-\Delta t}\mid \mathbf{x}_t,\mathbf{x}_0) \,\|\, p_{\theta}(\mathbf{x}_{t-\Delta t}\mid \mathbf{x}_t) \right),
    \label{eq:kl-local}
\end{equation}
which forms a local term in the negative evidence lower bound (NELBO).

SEDD~\citep{lou2023sedd} parameterizes $p_{\theta}(\mathbf{x}_{t-\Delta t}\mid\mathbf{x}_t)$ through rate ratios. Duo~\citep{sahoo2025duo} instead parameterizes the reverse transition as
$p_{\theta}(\mathbf{x}_{t-\Delta t}\mid\mathbf{x}_t)
=
p(\mathbf{x}_{t-\Delta t}\mid\mathbf{x}_t,\mathbf{x}_{\theta}(\mathbf{x}_t,t))$,
that is, the model directly predicts $\mathbf{x}_0$. Both approaches optimize the KL objective in Eq.~\eqref{eq:kl-local}. Let $m=\arg\max_j(\mathbf{x}_t)_j$ denote the current noisy token
and $y=\arg\max_j(\mathbf{x}_0)_j$ denote the clean token. Under the Duo parameterization, the token-level loss $\mathcal{L}_{\mathrm{Duo}}^{\ell}$ at position $\ell$ is
\begin{equation}
    \mathbb{E}_{t,\mathbf{x}_t} \frac{-\alpha_t'}{V\alpha_t}\! \left[\! \frac{V}{\bar{x}_{\theta,m}} \!-\! \frac{V}{\bar{x}_{0,m}} \!+\!
    \sum_{j=1}^{V}\! \frac{\bar{x}_{0,j}}{\bar{x}_{0,m}}\! \log \frac{ \bar{x}_{\theta,m}\bar{x}_{0,j} }{ \bar{x}_{\theta,j}\bar{x}_{0,m} }\!
    \right]\!.
    \label{eq:duo-loss}
\end{equation}
Summing over all positions $\ell$ gives the sentence-level training objective.

Appendix~\ref{app:sedd} shows that substituting the Duo $x_0$-parameterization into the uniform-state SEDD objective yields Eq.~\eqref{eq:duo-loss} exactly at the generator level.

SDDLM~\citep{zhu2025sddlm} follows the same parameterization as Duo, but introduces empirically simplified objectives for more stable and efficient training. The SDDLM and SDDLM-v1 losses are:
\begin{equation}
    \label{eq:sddlm-losses}
    \fitbox{0.98\textwidth}{$\displaystyle \mathcal{L}^{\ell}_{\mathrm{SD}} = -\mathbb{E}_{t,\mathbf{x}_t} \mathbb{I}_{\mathbf{x}_t\ne\mathbf{x}_0}
    \log x_{\theta,y}, \qquad \mathcal{L}^{\ell}_{\mathrm{SD1}} = -\mathbb{E}_{t,\mathbf{x}_t} \mathbb{I}_{\mathbf{x}_t\ne\mathbf{x}_0} \left[ \log
    x_{\theta,y} - \frac{1}{V}\sum_{j=1}^{V}\log x_{\theta,j} \right].$}
\end{equation}
Unlike $\mathcal{L}_{\mathrm{Duo}}$, which corresponds to a NELBO objective, these simplified losses are no longer NELBOs in theory, but have been observed to yield stronger empirical performance.

\section{Methodology}

In this section, we address the over-uniform and condition-target confusion that prevent UDLMs from scaling to complex tasks. The lack of a clean supervisory signal forces the model to rely on many denoising steps during sampling, yet complex tasks demand that simpler tokens be decoded quickly and serve as conditioning context. To resolve this tension, we first analyze the source of over-uniform and mitigate it with the LU loss (Sec.~\ref{sec:lu}), then introduce per-token time embeddings that provide token-level corruption hints (Sec.~\ref{sec:scale}). Together, the LU loss endows LUDI with stronger few-step generation capability, enabling it to rapidly decode high-confidence tokens and update their per-token time embeddings as conditions, thereby handling complex generation tasks effectively.

\subsection{Less Uniform Loss}
\label{sec:lu}
\paragraph{Over-uniform phenomenon.} In MDLMs, training based on ELBO has achieved widespread success, being used for large-scale pretraining or AR-to-diffusion training. The ELBO objective in UDLMs, while theoretically well grounded, is substantially less effective in practice. We attribute this to the \textit{mode-covering} property of the KL loss, which drives the model toward over-uniform and drowns out meaningful supervision.

To elucidate this issue, we simplify the complex original loss in Eq.\eqref{eq:duo-loss} into a CE-like form, as stated in the following proposition.

\begin{tcolorbox}[
    colback=gray!8,
    colframe=gray!25,
    boxrule=0.4pt,
    arc=1mm,
    left=1mm,
    right=1mm,
    top=1mm,
    bottom=1mm
]

\textbf{Proposition 1} (Proved in Appendix~\ref{app:proof-cor1}). Under mild assumptions detailed there, at a corrupted position $m\ne y$, the ELBO integrand, up to an additive term independent of $\theta$ and an $o(1)$ remainder as $V\to\infty$, is
\begingroup
\setlength{\fboxsep}{0.4pt}
\begin{equation}
    \fitbox{\linewidth}{$\displaystyle -\alpha_t'\!\left[ \frac{ \colorbox{blue!10!white}{$\log\bar{x}_{\theta,m}$}
    \colorbox{DarkGreen!12!white}{$-\log\bar{x}_{\theta,y}$} }{1-\alpha_t} \!-\!\frac{1}{V\alpha_t}\sum_{i=1}^{V}
    \colorbox{red!8!white}{$\log\bar{x}_{\theta,i}$} \right] = -\alpha_t'\!\left[
    \frac{\mathrm{CE}_{\mathrm{LS}}^{1-\alpha_t}(\bar{\mathbf{x}}_\theta,y)}{\alpha_t(1-\alpha_t)} +\frac{\log\bar{x}_{\theta,m}}{1-\alpha_t}
    \right].$}
\end{equation}
For already-clean positions $m=y$, if $x_{\theta,y}=\omega(1/V)$, the model-dependent contribution is $o(1)$ and therefore negligible as $V\to\infty$.
\endgroup
\end{tcolorbox}

When $\mathbf{x}_t^i \neq \mathbf{x}_0^i$, the loss decomposes into three terms:

$\qquad$(1) suppression of the current erroneous token \colorbox{blue!10!white}{$\log \bar{x}_{\theta,m}$}; $\quad$(2) learning of the correct token \colorbox{DarkGreen!12!white}{$-\log \bar{x}_{\theta,y}$};

$\qquad$(3) a smoothing term over the uniform distribution \colorbox{red!8!white}{$-\sum_i\log \bar{x}_{\theta,i}$}.

Notably, (2) and (3) can form a label-smoothed cross-entropy objective $\mathrm{CE}_{\mathrm{LS}}$ with effective target $\alpha_t\mathbf{x}_0+(1-\alpha_t)\mathbf{1}/V$. The smoothing strength $1-\alpha_t$ is unusually large. Under the common schedule $\alpha_t=1-t$ with $t\sim\mathcal{U}(0,1)$, on average only half of the probability mass is placed on clean token $\mathbf{x}_0$.

This stems from the mode-covering nature of the KL divergence. Whenever $p(\mathbf{x}_{t-\Delta t}\mid\mathbf{x}_t,\mathbf{x}_0)$ assigns probability to random tokens, the learned reverse transition kernel must cover those probabilities, otherwise it incurs a large penalty. Consequently, when $t$ is large, the model receives little effective supervision toward the clean token and is instead driven toward overly uniform predictions.

\textit{Remark 1.} This problem is circumvented in mask diffusion.
Although mask diffusion also employs a KL-based ELBO, the loss decouples into $t$-dependent and $t$-independent terms. In MDLMs, the probability of a token staying masked is computed analytically rather than learned.

\textit{Remark 2.} While SD and SDv1 in Eq.~\eqref{eq:sddlm-losses} heuristically remove the smoothing term \colorbox{red!8!white}{$-\sum_i\log \bar{x}_{\theta,i}$} and even introduce anti-smoothing, they also discard the suppression loss \colorbox{blue!10!white}{$\log \bar{x}_{\theta,m}$} on the erroneous token, thereby weakening the model's correction capability and lacking theoretical support.
\paragraph{LU loss.} To mitigate over-uniform, a simple fix is to directly remove term \colorbox{red!8!white}{$-\sum_i\log \bar{x}_{\theta,i}$}.
We obtain a \emph{less uniform} loss, termed the LU loss:
\begin{equation}
    \label{eq:lu-loss}
    \mathcal{L}_{\mathrm{LU}}^{\ell} \!=\! \mathbb{E}_{t,\mathbf{x}_t} \frac{-\alpha_t'}{1-\alpha_t} \left( \log \bar{x}_{\theta,m}\! -\! \log
    \bar{x}_{\theta,y} \right).
\end{equation}
Despite its simplicity, the LU loss has a direct reverse-process interpretation. At corrupted positions, it minimizes the finite model-dependent part of a Dirac-target KL.

\begin{tcolorbox}[
    colback=gray!8,
    colframe=gray!25,
    boxrule=0.4pt,
    arc=1mm,
    left=1mm,
    right=1mm,
    top=1mm,
    bottom=1mm
]
\textbf{Theorem 1} (Proved in Appendix~\ref{app:proof-cor2}).
At a corrupted position $m\ne y$, as $\Delta t\to0$,
\begin{equation}
    \mathrm{KL}\!\bigl(\delta(\mathbf{x}_0)\,\|\,p_\theta(\mathbf{x}_{t-\Delta t}\!\mid\!\mathbf{x}_t)\bigr) =-\log\!\frac{\alpha_{t-\Delta
    t}-\alpha_t}{V\alpha_{t-\Delta t}} +\log\bar{x}_{\theta,m}-\log\bar{x}_{\theta,y}+o(1).
\end{equation}
The first term is independent of $\theta$. Therefore, minimizing the finite model-dependent part of this KL is equivalent to minimizing $\mathcal{L}_{\mathrm{LU}}^{\ell}$.
\end{tcolorbox}

The LU loss simultaneously reduces the probability of the current incorrect token and increases the probability of the clean token $\mathbf{x}_0$. Moreover, the use of $\bar{\mathbf{x}}_\theta=V\alpha_t\mathbf{x}_\theta+(1-\alpha_t)\mathbf{1}$ prevents the logarithmic terms from becoming singular. As a result, the learned transition is consistently encouraged to move toward $\mathbf{x}_0$ for any time step. Intuitively, an ELBO-trained model behaves more like a \textit{diffusion model}, whereas the LU loss aligns more closely with a \textit{consistency model}, as illustrated in Figure~\ref{fig:ludi-overview}\subref{fig:ludi-loss}. Our LUDI loss offers two key advantages:

\begin{enumerate}
    \item LU loss avoids the excessive smoothing induced by the KL objective and instead aligns the parameterized transition rate with the clean token. This provides a stronger learning signal and improves few-step generation through consistency.
    \item Compared with the ELBO-based loss, our formulation involves only the two indices $y$ and $m$, making it computationally efficient. With optimized operators, LUDI can achieve a $4.39\times$ speedup and a $3.45\times$ memory reduction. The ELBO loss additionally applies a nonlinear transformation and reduction over all vocabulary entries, which is expensive for large vocabularies.
    (A detailed efficiency analysis is deferred to the Appendix~\ref{app:efficiency}.)
\end{enumerate}

\subsection{Scaling up UDLM}
\label{sec:scale}
\paragraph{Condition-target confusion.} When scaling UDLMs to the billion-parameter regime for challenging downstream tasks such as mathematics and code generation, a central obstacle is the ambiguity between conditioning context and denoising targets. Difficult reasoning tasks require the model to first establish a reliable condition, such as a mathematical equation, and then infer the target, such as the solution. In masked diffusion, this separation is explicit: clean tokens serve as conditions, while mask tokens identify the denoising targets. In UDLMs, however, the model is given only the global time $t$, which indicates the overall corruption level but not the status of each individual token. The model therefore struggles to distinguish condition tokens from target tokens and tends to denoise all positions simultaneously in a condition-independent manner. This behavior is particularly fatal for complex reasoning.

\paragraph{Per-token time embedding.} To address this ambiguity, we introduce per-token time embeddings. The denoiser is written as $\mathbf{x}_{\theta}(\mathbf{x}_t,\boldsymbol{\tau})$, where $\boldsymbol{\tau}=(\tau^1,\ldots,\tau^L)$ assigns a distinct time value to each position, as shown in Figure~\ref{fig:ludi-overview}\subref{fig:ludi-token-time}. Each $\tau^i$ indicates the probability that token $i$ has been randomly replaced, thereby providing the model with a token-level hint as to whether the token should be treated as condition or target.

To preserve the marginal distribution of $\mathbf{x}_t$, we use a hierarchical sampling procedure in the forward process. We first sample a global time $t\sim\mathcal{U}(0,1)$. Then, for each position $i$, we sample a per-token time $\tau^i\sim q_t$ and set the token to its clean value with probability $1-\tau^i$, or to a random vocabulary token with probability $\tau^i$. As long as $q_t$ has support on $[0,1]$ and satisfies $\mathbb{E}_{q_t}[\tau^i]=t$, the marginal distribution $p(\mathbf{x}_t\mid t)$ remains unchanged.

In practice, we choose $q_t$ to be a Beta distribution $\mathrm{Beta}(c t, c(1-t))$ with $c=2$, and inject the time signal via AdaLN~\citep{peebles2023dit}. Note that the per-token $\tau^i$ acts only as a hint; the LUDI loss itself still uses the global $t$.

\paragraph{Conditional sampling.} Benefiting from per-token time embeddings, LUDI naturally supports position-wise non-uniform conditional sampling. At initialization, non-prompt tokens are sampled uniformly from the vocabulary and assigned $\tau^i\sim\mathrm{Beta}(2,1)$, while prompt tokens are kept fixed with $\tau^i=0$. During denoising, non-prompt tokens are progressively updated until their token-level times $\tau^i$ approach zero.

Given the current token-level time $\tau^i_t$, the next value $\tau^i_{t-\Delta t}$ need not follow a uniform schedule and can instead be chosen adaptively for each position. For example, a high-confidence token can be decoded in a single step by setting $\tau^i_{t-\Delta t}=0$, after which it is treated as a conditioning token for subsequent generation. Consequently, LUDI can directly inherit samplers developed for MDLMs: each position either remains unchanged with $\tau^i_{t-\Delta t}=\tau^i_t$ or is committed by setting $\tau^i_{t-\Delta t}=0$. Empirically, this easy-to-hard denoising schedule is critical for complex reasoning tasks. We provide pseudocode for training and sampling in the Appendix~\ref{app:algorithm-details}.

\paragraph{AR to block-UDLM.} Training a large-scale UDLM from scratch is costly. We therefore initialize from an autoregressive (AR) model to maximally preserve acquired knowledge. To continue-training the AR model as a block-UDLM, we adopt two adaptation strategies.

\textit{Label shifting and complementary noise} (following Fast dLLMv2~\citep{wu2025fast}). Since an AR model predicts token $x_{i+1}$ from position $x_i$, we shift the labels left by one position. We also concatenate samples noised at $t$ and $1-t$ during training to reduce gradient variance.

\textit{Context-causal attention and AR loss} (following NB-Diff~\citep{tian2025nbdiff}).
We employ context-causal attention, where only the block currently being denoised receives bidirectional attention, while all preceding context retains causal attention. This design provides cleaner contextual signals for block generation and naturally enables inter-block KV-caching. We further regularize training with an auxiliary next-token prediction loss $\mathcal{L}_{\mathrm{AR}}$, which prevents the model from drifting excessively from its AR initialization. The overall objective is therefore $\mathcal{L}_{\mathrm{LUDI}} = \mathcal{L}_{\mathrm{LU}} + \lambda\,\mathcal{L}_{\mathrm{AR}}$, with $\lambda$ typically set to $0.5$ to supply supervision to the diffusion and AR components with comparable strength.

By combining these components, we present LUDI-7B, a 7B-scale UDLM capable of tackling complex reasoning tasks such as math and code generation, marking the transition of the UDLM paradigm from a conceptual idea to practical usability. Further design details are provided in the Appendix~\ref{app:block-udlm-mask}.

\section{Experiments}
We conduct experiments at two scales: small-scale pretraining (170M and 1B; Sec.~\ref{sec:small-scale}) and large-scale AR-to-UDLM continue-training (LUDI-7B; Sec.~\ref{sec:7b}). In the small-scale setting, where per-token time embeddings are not essential, we focus on the generation quality and training efficiency of the LU loss. In the large-scale setting, we evaluate LUDI-7B on complex reasoning tasks.

\subsection{Small-Scale LUDI}
\label{sec:small-scale}
\paragraph{Experimental setup.} For the 170M experiments, we train four UDLMs under the same configuration, using LU loss, Duo loss, SDDLM loss, and SDDLM-v1 loss, respectively. All models are trained on OpenWebText~\citep{Gokaslan2019OpenWeb} with a sequence length of 1024, using the GPT-2 tokenizer~\citep{Radford2019LanguageMA}. We use a batch size of 512 and a learning rate of $3\times10^{-4}$. For masked-diffusion baselines, we use pretrained RADD and MDLM models, trained on the same dataset for 400K and 1M steps, respectively. Generative perplexity is computed by GPT-2 Large on 512 samples. For the 1B experiments, we evaluate four UDLM models on downstream likelihood-based tasks. The models are trained on FineWeb~\citep{penedo2024fineweb} for 500K steps using the Llama
tokenizer~\citep{touvron2023llamaopenefficientfoundation}. The default sequence length is 2048, with 1\% variable-length data. We use a batch size of 256 and a learning rate of $2\times10^{-4}$. Additional experimental details are provided in Appendix~\ref{app:small-settings}.

\begin{figure}[t]
    \begin{minipage}[t]{0.40\textwidth}
        \vspace{0pt}
        \centering
        \includegraphics[width=0.98\linewidth]{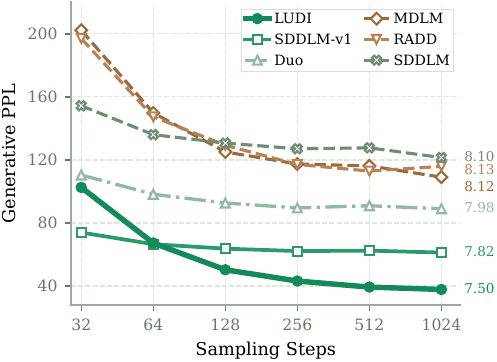}
        \caption{Gen.PPL from 32 to 1024 sampling steps.}
        \label{fig:small-scale-generation}
    \end{minipage}\hfill
    \begin{minipage}[t]{0.58\textwidth}
        \vspace{0pt}
        \centering
        \small
        \begingroup
        \renewcommand{\arraystretch}{1.1}
        \setlength{\tabcolsep}{0pt}
        \captionof{table}{Generation results (Gen.PPL and Entropy) at 200K and 500K checkpoints. All models sampled with 1024 steps. Lower Gen.PPL and higher entropy are better.}
        \label{tab:generation-results}
        \begin{tabularx}{\linewidth}{@{}l*{4}{>{\centering\arraybackslash}X}@{}}
        \toprule
        \multirow{2}{*}{\textbf{Model}}
        & \multicolumn{2}{c}{\textbf{200K Steps}}
        & \multicolumn{2}{c}{\textbf{500K Steps}} \\
        \cmidrule(lr){2-3} \cmidrule(lr){4-5}
        & \shortstack{\textbf{Gen.}\\ \textbf{PPL}} & \textbf{Entropy}
        & \shortstack{\textbf{Gen.}\\ \textbf{PPL}} & \textbf{Entropy} \\
        \midrule
        Duo       & 99.11 & 8.00 & 88.99 & 7.98 \\
        SDDLM     & 126.97 & 8.10 & 121.47 & 8.10 \\
        SDDLM-v1  & \second{64.22} & 7.82 & \second{61.18} & 7.82 \\
        \rowcolor{ludigreen} LUDI
                  & \best{37.09} & 7.53 & \best{37.76} & 7.50 \\
        \bottomrule
        \end{tabularx}
        \endgroup
    \end{minipage}
\end{figure}

\paragraph{Generative performance.} Figure~\ref{fig:small-scale-generation} compares generation quality under different numbers of sampling steps. With 1024 sampling steps, LUDI achieves a generative perplexity of 37.76 and an entropy of 7.50, outperforming both UDLM and MDLM baselines. Across most sampling budgets ($\geq64$ steps), LU loss consistently yields stronger generation quality. Although LUDI yields lower entropy, our analysis of Gen.PPL and diversity (see the Appendix~\ref{app:quality-diversity}) indicates that this reduction does not stem from mode collapse. UDLMs are often expected to enable few-step generation, but their quality can suffer from early saturation~\citep{deschenaux2026duo2}: increasing the number of sampling steps does not necessarily improve generation. Our experiments confirm this behavior. UDLMs outperform MDLMs in low-step regimes, yet SDDLM-v1 and Duo rapidly reach a plateau. In contrast, LUDI continues to improve as the sampling budget increases. We attribute this behavior to the stronger correction ability induced by LU loss. Compared with Duo loss, which contains a smoothing term, and SDDLM-style losses, which lack a correction term, LU loss more directly encourages the model to revise potentially erroneous tokens throughout sampling.

\begin{table}[t]
\centering
\small
\begingroup
\renewcommand{\arraystretch}{1.12}
\setlength{\tabcolsep}{3.5pt}
\caption{Zero-shot downstream evaluation of 1B models at 200k and 500k pretraining checkpoints. Full evaluation details are provided in Appendix~\ref{app:small-settings}.}
\label{tab:downstream-results}
\begin{tabular*}{\textwidth}{@{\extracolsep{\fill}}l>{\columncolor{ludigreen}}cccc>{\columncolor{ludigreen}}cccc@{}}
\toprule
\multirow{2}{*}{\textbf{Task}}
& \multicolumn{4}{c}{\textbf{200K Steps}}
& \multicolumn{4}{c}{\textbf{500K Steps}} \\
\cmidrule(lr){2-5}\cmidrule(lr){6-9}
& \textbf{LUDI} & \textbf{SDDLM} & \textbf{SDDLM-v1} & \textbf{Duo}
& \textbf{LUDI} & \textbf{SDDLM} & \textbf{SDDLM-v1} & \textbf{Duo} \\
\midrule
PIQA      & \best{60.12} & 57.40 & 55.33 & \second{58.11} & \best{62.19} & 61.15 & 60.28 & \second{61.37} \\
SIQA      & \best{36.13} & 35.11 & 34.70 & \second{35.67} & \second{36.54} & \best{37.26} & 35.47 & 36.03 \\
ARC-E     & \best{42.46} & \second{34.74} & 31.93 & 34.04 & \best{50.35} & \second{50.18} & 48.60 & 48.07 \\
HellaSwag & \best{34.42} & 30.85 & 30.05 & \second{31.82} & \best{35.31} & 34.22 & 33.31 & \second{35.28} \\
OBQA      & \best{26.60} & 25.80 & 24.80 & \second{26.40} & \second{28.60} & \best{29.80} & 27.20 & 28.00 \\
RACE      & \best{29.32} & 28.69 & 28.46 & \second{28.98} & \best{29.82} & 28.83 & 29.09 & \second{29.35} \\
Avg.      & \best{38.18} & 35.43 & 34.21 & \second{35.84} & \best{40.47} & \second{40.24} & 38.99 & 39.68 \\
\bottomrule
\end{tabular*}
\endgroup
\end{table}

\paragraph{Downstream task performance.} Table~\ref{tab:downstream-results} reports downstream results for 1B models on likelihood-based tasks, which probe the language modeling capability.
LUDI attains the highest average score and leads on most tasks, indicating both superior performance and stability. This advantage is consistent with the behavior of LU loss. By reducing confidence on erroneous samples and increasing confidence on correct ones, LUDI enlarges the relative margin of the correct answer.

\begin{table}[t]
\small
\centering
\setlength{\tabcolsep}{4pt}
\caption{LUDI-7B results and ablations. Metrics cover code generation (HumanEval, MBPP, HumanEval+, MBPP+), mathematical reasoning (GSM8K), instruction following (IFEval), knowledge-intensive question answering (MMLU, GPQA), and an overall average score (Avg.). Code benchmarks are evaluated using EvalPlus. The highest score in each column is marked in \best{bold}, and the second highest is \second{underlined}. AR $\to$ M/U denotes MDLM/UDLMs continue-trained from AR. In ablation, LUDI is based on Fast dLLM v2 (FD) and incorporates the LU loss, per-token time embedding, and NBDiff (ND) techniques. Full evaluation details are provided in
Appendix~\ref{app:7b-settings}.}
\label{tab:ludi7b-results}
\renewcommand{\arraystretch}{1.1}
\fitbox{\textwidth}{
\begin{tabular}{lccccccccccc}
\toprule
\multirow{2}{*}{\textbf{Model}} &
\multirow{2}{*}{\textbf{Size}} &
\multirow{2}{*}{\textbf{Type}} &
\multicolumn{2}{c}{\textbf{HumanEval}} &
\multicolumn{2}{c}{\textbf{MBPP}} &
\multirow{2}{*}{\textbf{GSM8K}} &
\multirow{2}{*}{\textbf{IFEval}} &
\multirow{2}{*}{\textbf{MMLU}} &
\multirow{2}{*}{\textbf{GPQA}} &
\multirow{2}{*}{\textbf{Avg.}} \\[-2pt]
\cmidrule(lr){4-5}\cmidrule(lr){6-7}
& & & Base $\quad$  & Plus $\quad$ & Base $\quad$ & Plus $\quad$
& & & & & \\
\hline
\multicolumn{12}{c}{\textit{Baselines}} \\
LLaDA        & 8B & MDLM & 35.4 & 31.7 & 31.5 & 28.6 & 78.6 & 59.9 & 65.5 & 31.8 & 45.4 \\
LLaDA-1.5    & 8B & MDLM & 52.4 & --   & 42.8 & --   & \second{83.3} & 58.2 & 66.0 & \best{36.9} & -- \\
LLaDA-MoE    & 7B & MDLM & 61.6 & --   & \second{70.0} & -- & 82.4 & 59.3 & 67.2 & -- & -- \\
Dream        & 7B & AR$\to$M & 57.9 & 53.7 & 68.3 & \second{56.1} & 81.0 & 62.5 & 67.0 & 33.0 & 59.9 \\
Qwen2.5-7B   & 7B & AR & 51.2 & 47.6 & 57.7 & 49.5 & 71.4 & \second{70.8} & \second{68.7} & 33.5 & 56.3 \\
Fast-dLLM v2 & 7B & AR$\to$M & \second{63.4} & \second{58.5} & 63.0 & 52.3 & \best{83.7} & 61.4 & 66.6 & 31.9 & \second{60.1} \\
\rowcolor{ludigreen}\textbf{LUDI-7B} & 7B & AR$\to$U & \best{68.1} & \best{63.4} & \best{71.5} & \best{60.7} & 75.2 & \best{78.0} & \best{71.1} & \second{34.3} & \best{65.3} \\
\hline
\multicolumn{12}{c}{\textit{Ablation }} \\

SD + \textsc{FDNB}  & 7B & AR$\to$U & 65.1 & 60.2 & 70.0 & 60.0 & 74.3 & 77.9 & \second{71.3} & 33.2 & 64.0 \\
SDv1 + \textsc{FDNB}  & 7B & AR$\to$U & \second{68.1} & 62.7 & 68.9 & 57.5 & 73.1 & \best{78.3} & \best{71.5} & 32.9 & 64.1 \\
LU + \textsc{FD}      & 7B & AR$\to$U & 59.1 & 54.9 & 67.7 & 56.3 & 64.7 & 75.1 & 70.8 & 32.4 & 60.1 \\
\textsc{FD}           & 7B & AR$\to$M & 62.2 & 58.5 & \second{72.0} & \best{61.1} & 71.3 & 77.7 & 71.2 & \second{33.3} & 63.4 \\
\textsc{FDNB}         & 7B & AR$\to$M & \best{68.3} & \second{62.8} & \best{72.2} & 59.8 & \best{76.0} & 77.2 & 70.4 & 31.0 & \second{64.7} \\
\rowcolor{ludigreen}\textbf{LUDI-7B} & 7B & AR$\to$U & \second{68.1} & \best{63.4} & 71.5 & \second{60.7} & \second{75.2} & \second{78.0} & 71.1 & \best{34.3} & \best{65.3} \\
\bottomrule
\end{tabular}
}
\end{table}

\paragraph{Training efficiency.} Table~\ref{tab:generation-results} and Table~\ref{tab:downstream-results} also compare models at 200K training steps. LUDI rapidly establishes clear advantages: the 170M model already exhibits stable generation, and the 1B LUDI outperforms all baselines across downstream tasks. We ascribe this to the cleaner learning objective provided by the LU loss. By avoiding the wasted supervision of label smoothing and directly matching $\mathbf{x}_0$, the LU loss offers a consistent signal: the model always learns to move away from erroneous tokens and toward the correct token. This improves training efficiency and data utilization, which is critical for large-scale training.

\subsection{LUDI-7B}
\label{sec:7b}
\paragraph{Experimental setup.}
We initialize our block UDLM from Qwen2.5-7B-Instruct~\cite{qwen2.5} and continue-train it on Dolci-Instruct-SFT~\cite{olmo2025olmo3}, a high-quality instruction dataset containing 2M samples. LUDI-7B is trained for 2500 steps with a batch size of 256 and a learning rate of $1\times 10^{-5}$, following the protocol of Fast dLLMv2~\cite{wu2025fast}. The block size is fixed to 32. In each transformer layer, we insert a learnable, zero-initialized AdaLN module before the multi-head self-attention and the MLP.
The AdaLN output at position $i$ is further multiplied by $\tau^i$. Consequently, the module is inactive wherever $\tau^i=0$, including prompt positions and tokens already committed during sampling. These AdaLN modules together with the time embeddings introduce only 78.7M additional parameters ($\sim$1\% of the total). During sampling, we adopt confidence-based parallel decoding.

\FloatBarrier
\begin{wrapfigure}[16]{tl}{0.49\textwidth}
    \vspace{-0.55\baselineskip}
    \centering
    \includegraphics[width=0.96\linewidth]{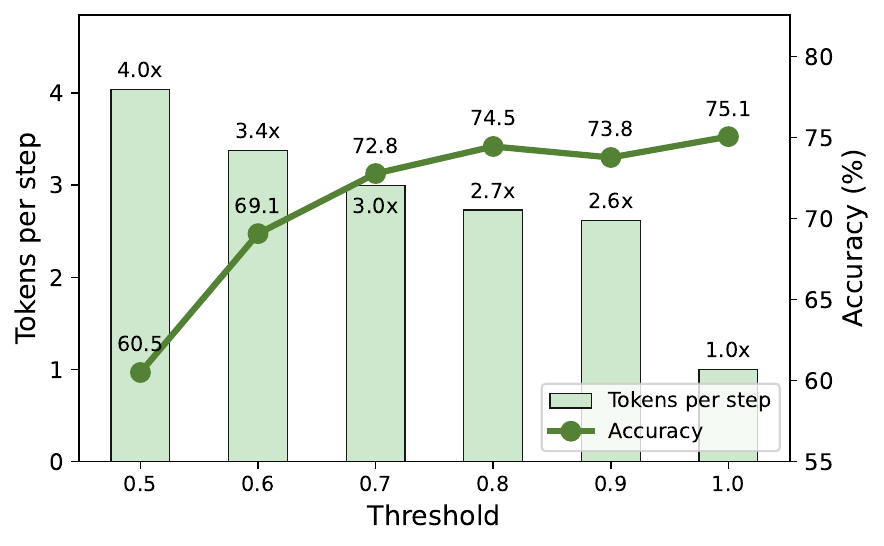}
    \caption{GSM8K performance and decoding speedup of LUDI-7B under different decoding thresholds.}
    \label{fig:ludi7b-speed-accuracy}
\end{wrapfigure}

\paragraph{Performance and speed.}
We compare LUDI with representative 7 to 8B autoregressive and diffusion models; the results are summarized in Table~\ref{tab:ludi7b-results}. LUDI-7B achieves an average score of 65.3, outperforming both the baselines and its AR initialization. It performs strongly on likelihood-based benchmarks (MMLU and GPQA) while also delivering competitive results on generation tasks. Although our method underperforms on mathematical reasoning, this mainly stems from the distributional bias of mathematical content in the training dataset. In Appendix~\ref{app:math-analysis}, we present additional evaluation results on mathematical reasoning and demonstrate that the mathematical reasoning performance of LUDI can be significantly improved through dataset curation.

With parallel decoding, LUDI-7B offers an attractive performance and efficiency trade-off.
By adjusting the confidence threshold, LUDI-7B maintains stable accuracy on GSM8K (Figure~\ref{fig:ludi7b-speed-accuracy}) for thresholds above 0.7, while yielding a 2.6 to 3.0 token-per-step speedup over AR decoding (1 token per step). These results confirm that UDLMs can handle complex generation tasks and possess substantial acceleration potential.

It is worth noting that the token-per-step metric represents a theoretical speedup; practical speedups are constrained by batch sampling and infrastructure overhead. In Appendix~\ref{app:wall-clock}, we report end-to-end efficiency comparisons across different batch sizes. At small batch sizes, we achieve actual speedups of 1.30--1.92$\times$. We acknowledge that achieving speedup at larger batch sizes requires further dLLM-specific infrastructure optimizations, which is orthogonal to our scope and is left as valuable future work.

\paragraph{Ablation study.}
To systematically examine the contribution of each design choice, we conduct controlled experiments on Dolci-Instruct under identical conditions (Table~\ref{tab:ludi7b-results}).
Our LUDI-7B builds upon Fast dLLM (FD) and integrates the LU loss, per-token time embeddings, and the tricks from NB-Diff (NB). (1) Without per-token time embeddings, the model suffers from severe condition-target confusion caused by position-uniform sampling, quickly collapsing to a trivial solution that repeats a single token. We therefore exclude this variant from comparison. (2) Replacing the LU loss with the SDv1 loss, which also mitigates over-uniform, still yields reasonable generations but with a clear performance drop. This suggests that complex generation requires explicit error correction: the model must be steered away from the current erroneous token. (3) The AR loss and causal context attention from NB-Diff prove important, indicating that block decoding in UDLMs benefits from higher-quality context.
(4) Compared to MDLM baselines, LUDI-7B surpasses FDNB trained under the same budget, demonstrating that UDLMs can profit from a more challenging training objective and may offer greater scaling potential under data-constrained regimes~\citep{vonrutte2025scaling}.

\section{Related Works}
\paragraph{Masked diffusion language models.}
MDLMs have become a competitive generative paradigm against AR models. Through large-scale pretraining~\citep{nie2025llada,bie2025llada2}, AR-to-diffusion adaptation~\citep{wu2025fastdllm,tian2025nbdiff,liu2025wedlm}, and reinforcement-learning post-training~\citep{zhao2025d1,zhong2026stabledrl,ni2026justgrpo}, MDLMs have achieved performance comparable to frontier LLMs on complex reasoning tasks while offering substantial inference-speed advantages. To better handle challenging problems, MDLM sampling has also moved beyond position-uniform unmasking toward strategic policies based on confidence~\citep{wu2025fastdllm} and entropy~\citep{benhamu2025ebsampler}. However, the binary mask/non-mask state space makes it difficult for MDLMs to correct already generated tokens. Moreover, the uninformative mask state naturally constrains parallel decoding~\citep{chen2026langflow}. Consequently, changing the masked generation paradigm has become an important direction for raising the ceiling of diffusion language models~\citep{ding2026did,vonrutte2025gidd}.
\paragraph{Uniform diffusion language models.}
UDLMs provide a complementary formulation where corrupted tokens are driven toward a uniform vocabulary distribution rather than an absorbing mask. This gives UDLMs potential advantages in few-step generation~\citep{sahoo2025duo,deschenaux2026duo2}, controllability~\citep{schiff2025guidance}, self-correction~\citep{schiff2026learnmistakes}, and scaling behavior under data- or compute-limited regimes~\citep{vonrutte2025scaling}. Nevertheless, existing analyses remain largely confined to small-scale settings or scaling-law studies, and UDLMs have not yet been scaled for complex reasoning tasks. Some works use uniform diffusion only as an auxiliary mechanism to improve the self-correction ability of MDLMs~\citep{schiff2026learnmistakes,vonrutte2025gidd}, or as a bridge toward continuous diffusion language modeling~\citep{sahoo2025duo,lee2026flm}. Efforts dedicated to improving UDLMs mainly focus on mitigating the sampling plateau with predictor-corrector samplers~\citep{deschenaux2026duo2} or simplifying the training objective~\citep{zhu2025sddlm}. However, these works either do not address, or only empirically touch upon, the over-uniform effect induced by UDLM training objectives; meanwhile, their samplers are not specifically designed for complex reasoning and largely remain position-uniform. These limitations create an inherent obstacle to scaling UDLMs, motivating our study.

\section{Conclusion}

This work presents LUDI as a step toward making UDLMs practical at scale. We show that the main barrier to scaling UDLMs lies not in the uniform noise prior itself, but in the excessive uniformity introduced by existing training objectives and sampling procedures. By introducing a less uniform loss and token-level time awareness, LUDI achieves cleaner supervision and a more efficient denoising process. Through LUDI-7B, we further demonstrate that UDLMs can be extended to complex reasoning tasks, narrowing the gap with AR and MDLM paradigms. These results suggest that UDLMs remain a promising yet underexplored direction.

\newpage
\section*{Limitations}

This work presents LUDI as an initial effort to scale uniform diffusion language models to 7B parameters and to apply them to complex reasoning tasks. Nevertheless, several limitations remain. (1) We have not yet fully activated the self-correction potential of UDLMs. The errors that arise during self-correction stem from the model's own imperfect generations, which are qualitatively different from the random noise introduced during training. Consequently, instilling this ability may require dedicated post-training stages that explicitly teach the model to detect and amend its own mistakes. (2) Due to computational resource constraints, we have not been able to pre-train a large-scale UDLM from scratch. Our approach of continue-training from an AR checkpoint efficiently preserves the original model's knowledge, yet it may also retain certain inductive biases of the autoregressive generation paradigm. We hope our work spurs future research into UDLMs pre-trained at scale from scratch, which may more fully exploit the unique advantages of the uniform diffusion framework.

\bibliographystyle{assets/plainnat}
\bibliography{references}

@inproceedings{sahoo2024mdlm,
  title     = {Simple and Effective Masked Diffusion Language Models},
  author    = {Sahoo, Subham Sekhar and Arriola, Marianne and Schiff, Yair and Gokaslan, Aaron and Marroquin, Edgar and Chiu, Justin T. and Rush, Alexander and Kuleshov, Volodymyr},
  booktitle = {Advances in Neural Information Processing Systems},
  year      = {2024}
}

@misc{nie2025llada,
  title         = {Large Language Diffusion Models},
  author        = {Nie, Shen and Zhu, Fengqi and You, Zebin and Zhang, Xiaolu and Ou, Jingyang and Hu, Jun and Zhou, Jun and Lin, Yankai and Wen, Ji-Rong and Li, Chongxuan},
  year          = {2025},
  eprint        = {2502.09992},
  archivePrefix = {arXiv},
  primaryClass  = {cs.CL}
}

@article{bie2025llada2,
  title={Llada2. 0: Scaling up diffusion language models to 100b},
  author={Bie, Tiwei and Cao, Maosong and Chen, Kun and Du, Lun and Gong, Mingliang and Gong, Zhuochen and Gu, Yanmei and Hu, Jiaqi and Huang, Zenan and Lan, Zhenzhong and Li, Chengxi and Li, Chongxuan and Li, Jianguo and Li, Zehuan and Liu, Huabin and Liu, Lin and Lu, Guoshan and Lu, Xiaocheng and Ma, Yuxin and Tan, Jianfeng and others},
  journal={arXiv preprint arXiv:2512.15745},
  year={2025}
}

@inproceedings{sahoo2025duo,
  title     = {The Diffusion Duality},
  author    = {Sahoo, Subham Sekhar and Deschenaux, Justin and Gokaslan, Aaron and Wang, Guanghan and Chiu, Justin and Kuleshov, Volodymyr},
  booktitle = {International Conference on Machine Learning},
  year      = {2025}
}

@article{zhu2025sddlm,
  title={Simple denoising diffusion language models},
  author={Zhu, Huaisheng and Chen, Zhengyu and Zhou, Shijie and Xie, Zhihui and Yuan, Yige and Chen, Shiqi and Guo, Zhimeng and Xu, Siyuan and Zhang, Hangfan and Honavar, Vasant and Xiao, Teng},
  journal={arXiv preprint arXiv:2510.22926},
  year={2025}
}

@misc{vonrutte2025scaling,
  title         = {Scaling Behavior of Discrete Diffusion Language Models},
  author        = {von R{\"u}tte, Dimitri and Fluri, Janis and Pooladzandi, Omead and Sch{\"o}lkopf, Bernhard and Hofmann, Thomas and Orvieto, Antonio},
  year          = {2025},
  eprint        = {2512.10858},
  archivePrefix = {arXiv},
  primaryClass  = {cs.LG}
}

@inproceedings{schiff2025guidance,
 author = {Schiff, Yair and Sahoo, Subham and Phung, Hao and Wang, Guanghan and Boshar, Sam and Dalla-torre, Hugo and Almeida, Bernardo and Rush, Alexander and Pierrot, Thomas and Kuleshov, Volodymyr},
 booktitle = {International Conference on Learning Representations},
 editor = {Y. Yue and A. Garg and N. Peng and F. Sha and R. Yu},
 pages = {43776--43821},
 title = {Simple Guidance Mechanisms for Discrete Diffusion Models},
 url = {https://proceedings.iclr.cc/paper_files/paper/2025/file/6cc31b44d88dce8380d36e81485cd07f-Paper-Conference.pdf},
 volume = {2025},
 year = {2025}
}

@inproceedings{ou2025radd,
  title={Your absorbing discrete diffusion secretly models the conditional distributions of clean data},
  author={Ou, Jingyang and Nie, Shen and Xue, Kaiwen and Zhu, Fengqi and Sun, Jiacheng and Li, Zhenguo and Li, Chongxuan},
  booktitle={International Conference on Learning Representations},
  volume={2025},
  pages={64972--65009},
  year={2025}
}

@article{lou2023sedd,
  title={Discrete diffusion modeling by estimating the ratios of the data distribution},
  author={Lou, Aaron and Meng, Chenlin and Ermon, Stefano},
  journal={arXiv preprint arXiv:2310.16834},
  year={2023}
}

@inproceedings{peebles2023dit,
  title     = {Scalable Diffusion Models with Transformers},
  author    = {William Peebles and Saining Xie},
  booktitle = {Proceedings of the IEEE/CVF International Conference on Computer Vision (ICCV)},
  year      = {2023},
  url       = {https://openaccess.thecvf.com/content/ICCV2023/html/Peebles_Scalable_Diffusion_Models_with_Transformers_ICCV_2023_paper.html}
}

@article{tian2025nbdiff,
  title={From next-token to next-block: A principled adaptation path for diffusion llms},
  author={Tian, Yuchuan and Liang, Yuchen and Zhang, Shuo and Shu, Yingte and Yang, Guangwen and He, Wei and Fang, Sibo and Guo, Tianyu and Han, Kai and Xu, Chao and Chen, Hanting and Chen, Xinghao and Wang, Yunhe},
  journal={arXiv preprint arXiv:2512.06776},
  year={2025}
}

@article{wu2025fast,
  title={Fast-dllm v2: Efficient block-diffusion llm},
  author={Wu, Chengyue and Zhang, Hao and Xue, Shuchen and Diao, Shizhe and Fu, Yonggan and Liu, Zhijian and Molchanov, Pavlo and Luo, Ping and Han, Song and Xie, Enze},
  journal={arXiv preprint arXiv:2509.26328},
  year={2025}
}

@misc{wu2025fastdllm,
  title         = {{Fast-dLLM}: Training-free Acceleration of Diffusion {LLM} by Enabling {KV} Cache and Parallel Decoding},
  author        = {Wu, Chengyue and Zhang, Hao and Xue, Shuchen and Liu, Zhijian and Diao, Shizhe and Zhu, Ligeng and Luo, Ping and Han, Song and Xie, Enze},
  year          = {2025},
  eprint        = {2505.22618},
  archivePrefix = {arXiv},
  primaryClass  = {cs.CL}
}

@misc{liu2025wedlm,
  title         = {{WeDLM}: Reconciling Diffusion Language Models with Standard Causal Attention for Fast Inference},
  author        = {Liu, Aiwei and He, Minghua and Zeng, Shaoxun and Zhang, Sijun and Zhang, Linhao and Wu, Chuhan and Jia, Wei and Liu, Yuan and Zhou, Xiao and Zhou, Jie},
  year          = {2025},
  eprint        = {2512.22737},
  archivePrefix = {arXiv},
  primaryClass  = {cs.CL}
}

@misc{zhao2025d1,
  title         = {d1: Scaling Reasoning in Diffusion Large Language Models via Reinforcement Learning},
  author        = {Zhao, Siyan and Gupta, Devaansh and Zheng, Qinqing and Grover, Aditya},
  year          = {2025},
  eprint        = {2504.12216},
  archivePrefix = {arXiv},
  primaryClass  = {cs.LG}
}

@misc{zhong2026stabledrl,
  title         = {Stabilizing Reinforcement Learning for Diffusion Language Models},
  author        = {Zhong, Jianyuan and Wang, Kaibo and Ding, Ding and Feng, Zijin and Bai, Haoli and Xiang, Yang and Sun, Jiacheng and Xu, Qiang},
  year          = {2026},
  eprint        = {2603.06743},
  archivePrefix = {arXiv},
  primaryClass  = {cs.LG}
}

@misc{ni2026justgrpo,
  title         = {The Flexibility Trap: Why Arbitrary Order Limits Reasoning Potential in Diffusion Language Models},
  author        = {Ni, Zanlin and Wang, Shenzhi and Yue, Yang and Yu, Tianyu and Zhao, Weilin and Hua, Yeguo and Chen, Tianyi and Song, Jun and Yu, Cheng and Zheng, Bo and Huang, Gao},
  year          = {2026},
  eprint        = {2601.15165},
  archivePrefix = {arXiv},
  primaryClass  = {cs.CL}
}

@misc{benhamu2025ebsampler,
  title         = {Accelerated Sampling from Masked Diffusion Models via Entropy Bounded Unmasking},
  author        = {Ben-Hamu, Heli and Gat, Itai and Severo, Daniel and Nolte, Niklas and Karrer, Brian},
  year          = {2025},
  eprint        = {2505.24857},
  archivePrefix = {arXiv},
  primaryClass  = {cs.LG}
}

@misc{chen2026langflow,
  title         = {{LangFlow}: Continuous Diffusion Rivals Discrete in Language Modeling},
  author        = {Chen, Yuxin and Liang, Chumeng and Sui, Hangke and Guo, Ruihan and Cheng, Chaoran and You, Jiaxuan and Liu, Ge},
  year          = {2026},
  eprint        = {2604.11748},
  archivePrefix = {arXiv},
  primaryClass  = {cs.CL}
}

@misc{ding2026did,
  title         = {Beyond Masks: Efficient, Flexible Diffusion Language Models via Deletion-Insertion Processes},
  author        = {Ding, Fangyu and Ding, Ding and Chen, Sijin and Wang, Kaibo and Xu, Peng and Feng, Zijin and Bai, Haoli and Han, Kai and Yan, Youliang and Yuan, Binhang and Sun, Jiacheng},
  year          = {2026},
  eprint        = {2603.23507},
  archivePrefix = {arXiv},
  primaryClass  = {cs.CL}
}

@misc{vonrutte2025gidd,
  title         = {Generalized Interpolating Discrete Diffusion},
  author        = {von R{\"u}tte, Dimitri and Fluri, Janis and Ding, Yuhui and Orvieto, Antonio and Sch{\"o}lkopf, Bernhard and Hofmann, Thomas},
  year          = {2025},
  eprint        = {2503.04482},
  archivePrefix = {arXiv},
  primaryClass  = {cs.LG}
}

@inproceedings{
  deschenaux2026duo2,
  title={The Diffusion Duality, Chapter {II}: $\Psi$-Samplers and Efficient Curriculum},
  author={Justin Deschenaux and Caglar Gulcehre and Subham Sekhar Sahoo},
  booktitle={The Fourteenth International Conference on Learning Representations},
  year={2026},
  url={https://openreview.net/forum?id=RSIoYWIzaP}
}

@article{lee2026flm,
  title={Flow map language models: One-step language modeling via continuous denoising},
  author={Lee, Chanhyuk and Yoo, Jaehoon and Agarwal, Manan and Shah, Sheel and Huang, Jerry and Raghunathan, Aditi and Hong, Seunghoon and Boffi, Nicholas M and Kim, Jinwoo},
  journal={arXiv preprint arXiv:2602.16813},
  year={2026}
}

@misc{schiff2026learnmistakes,
  title         = {Learn from Your Mistakes: Self-Correcting Masked Diffusion Models},
  author        = {Schiff, Yair and Belhasin, Omer and Uziel, Roy and Wang, Guanghan and Arriola, Marianne and Turok, Gilad and Elad, Michael and Kuleshov, Volodymyr},
  year          = {2026},
  eprint        = {2602.11590},
  archivePrefix = {arXiv},
  primaryClass  = {cs.LG}
}

@misc{qwen2.5,
    title = {Qwen2.5: A Party of Foundation Models},
    url = {https://qwenlm.github.io/blog/qwen2.5/},
    author = {Qwen Team},
    month = {September},
    year = {2024}
}

@misc{olmo2025olmo3,
title={Olmo 3},
author={Team Olmo and Allyson Ettinger and Amanda Bertsch and Bailey Kuehl and David Graham and David Heineman and Dirk Groeneveld and Faeze Brahman and Finbarr Timbers and Hamish Ivison and Jacob Morrison and Jake Poznanski and Kyle Lo and Luca Soldaini and Matt Jordan and Mayee Chen and Michael Noukhovitch and Nathan Lambert and Pete Walsh and Pradeep Dasigi and Robert Berry and Saumya Malik and Saurabh Shah and Scott Geng and Shane Arora and Shashank Gupta and Taira Anderson and Teng Xiao and Tyler Murray and Tyler Romero and Victoria Graf and Akari Asai and Akshita Bhagia and Alexander Wettig and Alisa Liu and Aman Rangapur and Chloe Anastasiades and Costa Huang and Dustin Schwenk and Harsh Trivedi and Ian Magnusson and Jaron Lochner and Jiacheng Liu and Lester James V. Miranda and Maarten Sap and Malia Morgan and Michael Schmitz and Michal Guerquin and Michael Wilson and Regan Huff and Ronan Le Bras and Rui Xin and Rulin Shao and Sam Skjonsberg and Shannon Zejiang Shen and Shuyue Stella Li and Tucker Wilde and Valentina Pyatkin and Will Merrill and Yapei Chang and Yuling Gu and Zhiyuan Zeng and Ashish Sabharwal and Luke Zettlemoyer and Pang Wei Koh and Ali Farhadi and Noah A. Smith and Hannaneh Hajishirzi},
year={2025},
eprint={2512.13961},
archivePrefix={arXiv},
primaryClass={cs.CL},
url={https://arxiv.org/abs/2512.13961},
}

@misc{Gokaslan2019OpenWeb,
    title={OpenWebText Corpus},
    author={Gokaslan, Aaron and Cohen, Vanya and Pavlick, Ellie and Tellex, Stefanie},
    howpublished={\url{http://Skylion007.github.io/OpenWebTextCorpus}},
    year={2019}
}

@article{penedo2024fineweb,
  title={The fineweb datasets: Decanting the web for the finest text data at scale},
  author={Penedo, Guilherme and Kydl{\'\i}{\v{c}}ek, Hynek and {Ben allal}, Loubna and Lozhkov, Anton and Mitchell, Margaret and Raffel, Colin and Von Werra, Leandro and Wolf, Thomas},
  journal={Advances in Neural Information Processing Systems},
  volume={37},
  pages={30811--30849},
  year={2024}
}

@misc{Radford2019LanguageMA,
  title={Language Models are Unsupervised Multitask Learners},
  author={Alec Radford and Jeff Wu and Rewon Child and David Luan and Dario Amodei and Ilya Sutskever},
  year={2019},
  url={https://api.semanticscholar.org/CorpusID:160025533}
}

@misc{touvron2023llamaopenefficientfoundation,
      title={LLaMA: Open and Efficient Foundation Language Models}, 
      author={Hugo Touvron and Thibaut Lavril and Gautier Izacard and Xavier Martinet and Marie-Anne Lachaux and Timothée Lacroix and Baptiste Rozière and Naman Goyal and Eric Hambro and Faisal Azhar and Aurelien Rodriguez and Armand Joulin and Edouard Grave and Guillaume Lample},
      year={2023},
      eprint={2302.13971},
      archivePrefix={arXiv},
      primaryClass={cs.CL},
      url={https://arxiv.org/abs/2302.13971}, 
}

@article{ye2025dream,
  title={Dream 7b: Diffusion large language models},
  author={Ye, Jiacheng and Xie, Zhihui and Zheng, Lin and Gao, Jiahui and Wu, Zirui and Jiang, Xin and Li, Zhenguo and Kong, Lingpeng},
  journal={arXiv preprint arXiv:2508.15487},
  year={2025}
}

@article{llada1.5,
  title={Llada 1.5: Variance-reduced preference optimization for large language diffusion models},
  author={Zhu, Fengqi and Wang, Rongzhen and Nie, Shen and Zhang, Xiaolu and Wu, Chunwei and Hu, Jun and Zhou, Jun and Chen, Jianfei and Lin, Yankai and Wen, Ji-Rong and Li, Chongxuan},
  journal={arXiv preprint arXiv:2505.19223},
  year={2025}
}

@article{lladamoe,
  title={Llada-moe: A sparse moe diffusion language model},
  author={Zhu, Fengqi and You, Zebin and Xing, Yipeng and Huang, Zenan and Liu, Lin and Zhuang, Yihong and Lu, Guoshan and Wang, Kangyu and Wang, Xudong and Wei, Lanning and Guo, Hongrui and Hu, Jiaqi and Ye, Wentao and Chen, Tieyuan and Li, Chenchen and Tang, Chengfu and Feng, Haibo and Hu, Jun and Zhou, Jun and Zhang, Xiaolu and others},
  journal={arXiv preprint arXiv:2509.24389},
  year={2025}
}

@article{evalplus,
  title={Is your code generated by chatgpt really correct? rigorous evaluation of large language models for code generation},
  author={Liu, Jiawei and Xia, Chunqiu Steven and Wang, Yuyao and Zhang, Lingming},
  journal={Advances in neural information processing systems},
  volume={36},
  pages={21558--21572},
  year={2023}
}

@article{humaneval,
  title={Evaluating large language models trained on code},
  author={Chen, Mark and Tworek, Jerry and Jun, Heewoo and Yuan, Qiming and Pinto, Henrique Ponde De Oliveira and Kaplan, Jared and Edwards, Harri and Burda, Yuri and Joseph, Nicholas and Brockman, Greg and Ray, Alex and Puri, Raul and Krueger, Gretchen and Petrov, Michael and Khlaaf, Heidy and Sastry, Girish and Mishkin, Pamela and Chan, Brooke and Gray, Scott and Ryder, Nick and others},
  journal={arXiv preprint arXiv:2107.03374},
  year={2021}
}

@article{mbpp,
  title={Program synthesis with large language models},
  author={Austin, Jacob and Odena, Augustus and Nye, Maxwell and Bosma, Maarten and Michalewski, Henryk and Dohan, David and Jiang, Ellen and Cai, Carrie and Terry, Michael and Le, Quoc and Sutton, Charles},
  journal={arXiv preprint arXiv:2108.07732},
  year={2021}
}

@article{gsm,
  title={Training verifiers to solve math word problems},
  author={Cobbe, Karl and Kosaraju, Vineet and Bavarian, Mohammad and Chen, Mark and Jun, Heewoo and Kaiser, Lukasz and Plappert, Matthias and Tworek, Jerry and Hilton, Jacob and Nakano, Reiichiro and Hesse, Christopher and Schulman, John},
  journal={arXiv preprint arXiv:2110.14168},
  year={2021}
}

@article{ifeval,
  title={Instruction-following evaluation for large language models},
  author={Zhou, Jeffrey and Lu, Tianjian and Mishra, Swaroop and Brahma, Siddhartha and Basu, Sujoy and Luan, Yi and Zhou, Denny and Hou, Le},
  journal={arXiv preprint arXiv:2311.07911},
  year={2023}
}

@article{rein2023gpqa,
  title={Gpqa: A graduate-level google-proof q\&a benchmark},
  author={Rein, David and Hou, Betty Li and Stickland, Asa Cooper and Petty, Jackson and Pang, Richard Yuanzhe and Dirani, Julien and Michael, Julian and Bowman, Samuel R},
  journal={arXiv preprint arXiv:2311.12022},
  year={2023}
}

@article{mmlu,
  title={Measuring massive multitask language understanding},
  author={Hendrycks, Dan and Burns, Collin and Basart, Steven and Zou, Andy and Mazeika, Mantas and Song, Dawn and Steinhardt, Jacob},
  journal={arXiv preprint arXiv:2009.03300},
  year={2020}
}

@inproceedings{bisk2020piqa,
  title={Piqa: Reasoning about physical commonsense in natural language},
  author={Bisk, Yonatan and Zellers, Rowan and Le Bras, Ronan and Gao, Jianfeng and Choi, Yejin},
  booktitle={Proceedings of the AAAI conference on artificial intelligence},
  volume={34},
  pages={7432--7439},
  year={2020}
}

@inproceedings{sap-etal-2019-social,
    title = "Social {IQ}a: Commonsense Reasoning about Social Interactions",
    author = "Sap, Maarten  and
      Rashkin, Hannah  and
      Chen, Derek  and
      Le Bras, Ronan  and
      Choi, Yejin",
    editor = "Inui, Kentaro  and
      Jiang, Jing  and
      Ng, Vincent  and
      Wan, Xiaojun",
    booktitle = "Proceedings of the 2019 Conference on Empirical Methods in Natural Language Processing and the 9th International Joint Conference on Natural Language Processing (EMNLP-IJCNLP)",
    month = nov,
    year = "2019",
    address = "Hong Kong, China",
    publisher = "Association for Computational Linguistics",
    url = "https://aclanthology.org/D19-1454/",
    doi = "10.18653/v1/D19-1454",
    pages = "4463--4473"
}

@article{clark2018think,
  title={Think you have solved question answering? try arc, the ai2 reasoning challenge},
  author={Clark, Peter and Cowhey, Isaac and Etzioni, Oren and Khot, Tushar and Sabharwal, Ashish and Schoenick, Carissa and Tafjord, Oyvind},
  journal={arXiv preprint arXiv:1803.05457},
  year={2018}
}

@inproceedings{zellers-etal-2019-hellaswag,
    title = "{H}ella{S}wag: Can a Machine Really Finish Your Sentence?",
    author = "Zellers, Rowan  and
      Holtzman, Ari  and
      Bisk, Yonatan  and
      Farhadi, Ali  and
      Choi, Yejin",
    editor = "Korhonen, Anna  and
      Traum, David  and
      M{\`a}rquez, Llu{\'i}s",
    booktitle = "Proceedings of the 57th Annual Meeting of the Association for Computational Linguistics",
    month = jul,
    year = "2019",
    address = "Florence, Italy",
    publisher = "Association for Computational Linguistics",
    url = "https://aclanthology.org/P19-1472/",
    doi = "10.18653/v1/P19-1472",
    pages = "4791--4800"
}

@inproceedings{mihaylov-etal-2018-suit,
    title = "Can a Suit of Armor Conduct Electricity? A New Dataset for Open Book Question Answering",
    author = "Mihaylov, Todor  and
      Clark, Peter  and
      Khot, Tushar  and
      Sabharwal, Ashish",
    editor = "Riloff, Ellen  and
      Chiang, David  and
      Hockenmaier, Julia  and
      Tsujii, Jun{'}ichi",
    booktitle = "Proceedings of the 2018 Conference on Empirical Methods in Natural Language Processing",
    month = oct # "-" # nov,
    year = "2018",
    address = "Brussels, Belgium",
    publisher = "Association for Computational Linguistics",
    url = "https://aclanthology.org/D18-1260/",
    doi = "10.18653/v1/D18-1260",
    pages = "2381--2391"
}

@inproceedings{lai-etal-2017-race,
    title = "{RACE}: Large-scale {R}e{A}ding Comprehension Dataset From Examinations",
    author = "Lai, Guokun  and
      Xie, Qizhe  and
      Liu, Hanxiao  and
      Yang, Yiming  and
      Hovy, Eduard",
    editor = "Palmer, Martha  and
      Hwa, Rebecca  and
      Riedel, Sebastian",
    booktitle = "Proceedings of the 2017 Conference on Empirical Methods in Natural Language Processing",
    month = sep,
    year = "2017",
    address = "Copenhagen, Denmark",
    publisher = "Association for Computational Linguistics",
    url = "https://aclanthology.org/D17-1082/",
    doi = "10.18653/v1/D17-1082",
    pages = "785--794"
}

@inproceedings{nie2025scaling,
  title={Scaling up masked diffusion models on text},
  author={Nie, Shen and Zhu, Fengqi and Du, Chao and Pang, Tianyu and Liu, Qian and Zeng, Guangtao and Lin, Min and Li, Chongxuan},
  booktitle={International Conference on Learning Representations},
  volume={2025},
  pages={82974--82997},
  year={2025}
}

@misc{NemotronPostTrainingDatasetV2,
      author = {Nathawani, Dhruv and Ding, Shuoyang and Lavrukhin, Vitaly and Gitman, Igor and Majumdar, Somshubra and Bakhturina, Evelina and Ginsburg, Boris and Polak Scowcroft, Jane},
      title = {{Nemotron-Post-Training-Dataset-v2}},
      version = {2.0},
      publisher = {{NVIDIA}},
      year = {2025}, month = aug,
      url = {https://huggingface.co/datasets/nvidia/Nemotron-Post-Training-Dataset-v2}
}

@inproceedings{lightman2024let,
  title={Let's verify step by step},
  author={Lightman, Hunter and Kosaraju, Vineet and Burda, Yuri and Edwards, Harrison and Baker, Bowen and Lee, Teddy and Leike, Jan and Schulman, John and Sutskever, Ilya and Cobbe, Karl},
  booktitle={International Conference on Learning Representations},
  volume={2024},
  pages={39578--39601},
  year={2024}
}

\appendix

\providecommand{\mcV}{\mathcal{V}}
\providecommand{\Vsize}{V}
\providecommand{\Cat}{\mathrm{Cat}}
\providecommand{\KL}{\mathrm{KL}}
\providecommand{\E}{\mathbb{E}}
\providecommand{\R}{\mathbb{R}}
\providecommand{\1}{\mathbf{1}}
\providecommand{\I}{\mathbb{I}}
\providecommand{\vx}[1]{\bm{#1}}
\providecommand{\xzero}{\vx{x}_0}
\providecommand{\xt}{\vx{x}_t}
\providecommand{\xprev}{\vx{x}_{t-\Delta t}}
\providecommand{\xth}{\vx{x}_{\theta}}
\providecommand{\Loss}{\mathcal{L}}
\providecommand{\alphat}{\alpha_t}
\providecommand{\alphap}{\alpha_t'}
\providecommand{\mask}{\mathtt{m}}

\clearpage
\begingroup
\setlength{\abovedisplayskip}{5pt plus 1pt minus 1pt}
\setlength{\belowdisplayskip}{5pt plus 1pt minus 1pt}
\setlength{\abovedisplayshortskip}{3pt plus 1pt minus 1pt}
\setlength{\belowdisplayshortskip}{3pt plus 1pt minus 1pt}

\section{Proofs and Derivations}

We prove the results of the main text for one token; sequence-level objectives
follow by summing over positions.  Let $\mcV=\{1,\ldots,V\}$, let $\vx e_i$ be
the one-hot vector for token $i$, and write $\vx x_0=\vx e_y$ and
$\vx x_t=\vx e_m$.  Thus $m\ne y$ denotes a corrupted position.

Assume that $t\mapsto\alpha_t$ is continuous and non-increasing on $[0,1]$,
continuously differentiable on $[0,1)$, and satisfies $\alpha_0=1$,
$\alpha_t>0$ for $t<1$, and $\alpha_1=0$.  For
$\Delta^{V-1}:=\{\vx x\in\R_{\geq0}^{V}:\sum_i x_i=1\}$, define
\begin{equation}
    \bar{\vx x}:=V\alpha_t\vx x+(1-\alpha_t)\1, \qquad \bar{x}_i=V\alpha_t x_i+1-\alpha_t.
\end{equation}
Then $\sum_i\bar{x}_i=V$, so $\bar{\vx x}/V$ is a probability vector.  We use
$\bar{x}_{0,i}$ and $\bar{x}_{\theta,i}$ for the vectors obtained from
$\vx x_0$ and $\vx x_\theta(\vx x_t,t)$, respectively.

\subsection{Uniform-State Diffusion Preliminaries}

We first verify the forward law and reverse posterior, then connect the SEDD and
Duo objectives.

\subsubsection{Forward Process}

We verify that the stated rate matrix generates the uniform-state marginal.

For $t\in[0,1)$, the uniform-state transition rate matrix is
\begin{equation}
    Q_t(i,j)=-\frac{\alpha_t'}{\alpha_t}\left(\frac{1}{V}-\I_{i=j}\right).
\end{equation}
Since $\alpha_t'\leq0$, the off-diagonal entries are nonnegative, and each row
sums to zero.  Starting from $\vx x_0=\vx e_y$, consider
\begin{equation}
    p_t(j\mid y):=\alpha_t\I_{j=y}+(1-\alpha_t)\frac{1}{V}.
    \label{eq:app-uniform-marginal-scalar}
\end{equation}
Direct substitution gives
\begin{equation}
    \frac{d}{dt}p_t(j\mid y)=\alpha_t'\left(\I_{j=y}-\frac{1}{V}\right)=\bigl[p_t(\cdot\mid y)Q_t\bigr]_j.
\end{equation}
Together with $p_0(j\mid y)=\I_{j=y}$, uniqueness of the forward equation yields
\begin{equation}
    p(\vx x_t\mid \vx x_0)=\Cat\left(\cdot;\,\alpha_t\vx x_0+(1-\alpha_t)\frac{\1}{V}\right).
\end{equation}
\subsubsection{Exact Reverse Posterior}
\label{app:backward-posteriors}

We obtain the reverse posterior by applying Bayes' rule to the forward
transitions.

Let $0\le s<t\le1$ and $\alpha_{t\mid s}:=\alpha_t/\alpha_s$.  The transition
from $s$ to $t$ is
\begin{equation}
    p(\vx x_t=\vx e_m\mid \vx x_s=\vx e_k)=\alpha_{t\mid s}\I_{m=k}+\frac{1-\alpha_{t\mid s}}{V}.
\end{equation}
For an observation $\vx x_t=\vx e_m$ of positive probability, Bayes' rule gives
\begin{equation}
    \fitbox{0.98\textwidth}{$\displaystyle p(\vx x_s=\vx e_k\mid \vx x_t=\vx e_m,\vx x_0=\vx e_y) =\frac{\left(\alpha_{t\mid
    s}\I_{m=k}+\frac{1-\alpha_{t\mid s}}{V}\right)\left(\alpha_s\I_{k=y}+\frac{1-\alpha_s}{V}\right)}{\alpha_t\I_{m=y}+\frac{1-\alpha_t}{V}}.$}
\end{equation}

Let $\vx\pi_{t\to s}\in\Delta^{V-1}$ denote the categorical parameter vector
of this posterior, with
\begin{equation}
    [\vx\pi_{t\to s}]_k :=p(\vx x_s=\vx e_k\mid \vx x_t=\vx e_m,\vx x_0=\vx e_y).
\end{equation}

Multiplying numerator and denominator by $V$ and using
$\alpha_{t\mid s}\alpha_s=\alpha_t$, the numerator of coordinate $k$ is
\begin{align}
    &V\left(\alpha_{t\mid s}\I_{m=k}+\frac{1-\alpha_{t\mid s}}{V}\right)\left(\alpha_s\I_{k=y}+\frac{1-\alpha_s}{V}\right) \notag\\
     &\quad=V\alpha_t\I_{m=k}\I_{k=y}+(\alpha_{t\mid s}-\alpha_t)\I_{m=k}+(\alpha_s-\alpha_t)\I_{k=y}+\frac{(1-\alpha_{t\mid s})(1-\alpha_s)}{V}.
\end{align}
Collecting these coordinates yields
\begin{align}
    \vx \pi_{t\to s} &=\frac{1}{V\alpha_t\I_{m=y}+1-\alpha_t} \Bigg[ V\alpha_t\,\vx x_t\odot\vx x_0 +(\alpha_{t\mid s}-\alpha_t)\vx x_t
    +(\alpha_s-\alpha_t)\vx x_0 +(1-\alpha_{t\mid s})(1-\alpha_s)\frac{\1}{V} \Bigg],
    \label{eq:app-uniform-posterior-vector}
\end{align}

\subsubsection{SEDD--Duo Reparameterization}
\label{app:sedd}

We show that the SEDD score-entropy objective reduces exactly to the Duo
objective under the $x_0$-parameterization.

For $0\leq s<t<1$, consider the local reverse KL between the exact posterior
and the learned reverse transition,
\begin{equation}
    \KL\!\bigl( p(\vx x_s\mid \vx x_t,\vx x_0) \;\|\; p_\theta(\vx x_s\mid \vx x_t) \bigr).
\end{equation}
The posterior entropy is independent of $\theta$, so only the cross-entropy is
model-dependent.  Fix $\vx x_0=\vx e_y$.  The conditional SEDD target is
\begin{equation}
    s^\star(j,m,t)=\frac{p_t(\vx e_j\mid \vx x_0)}{p_t(\vx e_m\mid \vx x_0)}=\frac{\bar{x}_{0,j}}{\bar{x}_{0,m}}.
\end{equation}
The ratio is evaluated only when the denominator is positive, with
$0\log0=0$.  The token-level denoising score-entropy objective is
\begin{equation}
    \Loss_{\mathrm{DSE}}^{\ell} = \E_{t,\vx x_t}\!\sum_{j\ne m} Q_t(j,m)\! \bigl[ s_\theta(j,m,t)-s^\star(j,m,t)\log s_\theta(j,m,t)
    +s^\star(j,m,t)\bigl(\log s^\star(j,m,t)-1\bigr) \bigr].
\end{equation}
For $j\ne m$, $Q_t(j,m)=-\alpha_t'/(V\alpha_t)$.  The omitted $j=m$ summand is
zero and may be restored.  Under the Duo parameterization,
\begin{equation}
    s_\theta(j,m,t)=\frac{\bar{x}_{\theta,j}}{\bar{x}_{\theta,m}}.
\end{equation}
Using $\sum_j\bar{x}_{\theta,j}=\sum_j\bar{x}_{0,j}=V$,
\begin{align}
    \Loss_{\mathrm{DSE}}^{\ell} &=\E_{t,\vx x_t}\frac{-\alpha_t'}{V\alpha_t}\left[ \sum_{j=1}^{V}\frac{\bar{x}_{\theta,j}}{\bar{x}_{\theta,m}}
    -\sum_{j=1}^{V}\frac{\bar{x}_{0,j}}{\bar{x}_{0,m}} +\sum_{j=1}^{V}\frac{\bar{x}_{0,j}}{\bar{x}_{0,m}}
    \log\frac{\bar{x}_{\theta,m}\bar{x}_{0,j}}{\bar{x}_{\theta,j}\bar{x}_{0,m}} \right] \notag\\
     &=\E_{t,\vx x_t}\frac{-\alpha_t'}{V\alpha_t}\left[ \frac{V}{\bar{x}_{\theta,m}}-\frac{V}{\bar{x}_{0,m}}
    +\sum_{j=1}^{V}\frac{\bar{x}_{0,j}}{\bar{x}_{0,m}} \log\frac{\bar{x}_{\theta,m}\bar{x}_{0,j}}{\bar{x}_{\theta,j}\bar{x}_{0,m}} \right].
    \label{eq:app-duo-weighted}
\end{align}
This is Eq.~\eqref{eq:duo-loss}.

\subsubsection{Contrast with Mask Diffusion}

We contrast the uniform-state objective with the simpler posterior structure of
absorbing-state mask diffusion.

For comparison, standard absorbing-state mask diffusion on
$\mcV\cup\{\mathtt{[MASK]}\}$ has the forward law \citep{sahoo2024mdlm}
\begin{equation}
    p_t^{\mathrm M}(y\mid y)=\alpha_t, \qquad p_t^{\mathrm M}(\mathtt{[MASK]}\mid y)=1-\alpha_t,
\end{equation}
Conditioning on $x_t=\mathtt{[MASK]}$, the two possible states at time $s$ have
joint probabilities $\alpha_s-\alpha_t$ and $1-\alpha_s$.  Since
$p(x_t=\mathtt{[MASK]}\mid x_0=y)=1-\alpha_t$, Bayes' rule gives
\begin{equation}
    a_{t,s}=\frac{\alpha_s-\alpha_t}{1-\alpha_t}, \qquad b_{t,s}=\frac{1-\alpha_s}{1-\alpha_t}.
\end{equation}
Under the $x_0$-parameterization, the learned posterior assigns
$a_{t,s}x_{\theta,j}$ to token $j$ and $b_{t,s}$ to $\mathtt{[MASK]}$.  Hence
\begin{equation}
    \KL\!\left( p(x_s\mid x_t=\mathtt{[MASK]},x_0=y) \,\|\, p_\theta(x_s\mid x_t=\mathtt{[MASK]}) \right) =-a_{t,s}\log x_{\theta,y}.
\end{equation}
Thus the model learns only the clean-token cross-entropy; both posterior
weights are analytic.

\subsection{ELBO Loss and LU Loss}

This subsection proves the two claims that motivate the LU loss.  Proposition~1
separates the uniform-state ELBO into interpretable terms, and Theorem~1 shows
that the LU contrast is the finite model-dependent part of a small-step
Dirac-target KL.

\subsubsection{Proof of Proposition 1}
\label{app:proof-cor1}

We first state the assumptions precisely.  The proof then treats corrupted and
already-clean positions separately, because their clean noisy-vector
coordinates have different forms.

\paragraph{Proposition 1 (Restated).}
Fix $0<\varepsilon<1/2$ and restrict attention to times satisfying
$\alpha_t\in[\varepsilon,1-\varepsilon]$, on which $|\alpha_t'|$ is bounded.
Suppose that, uniformly over these times and corrupted positions $m\ne y$,
$x_{\theta,m}=1/V+o(1/V)$ as $V\to\infty$.  After omitting additive terms
independent of $\theta$, the token-level ELBO integrand is
\begin{equation}
    -\alpha_t'\!\left[ \frac{\log\bar{x}_{\theta,m}-\log\bar{x}_{\theta,y}}{1-\alpha_t} -\frac{1}{V\alpha_t}\sum_{i=1}^{V}\log\bar{x}_{\theta,i}
    \right]+o(1),
\end{equation}
where the error is uniform over the stated region and therefore remains
$o(1)$ after averaging.  If, in addition,
$x_{\theta,y}=\omega(1/V)$ uniformly at already-clean positions $m=y$, their
model-dependent contribution is also uniformly $o(1)$ and remains $o(1)$
after averaging.

\paragraph{Proof.}
We first isolate the model-dependent part at corrupted positions, then bound
the contribution from already-clean positions.  All calculations are
pointwise in $(\vx x_0,t,\vx x_t)$; additive terms independent of $\theta$ are
omitted.

\paragraph{Corrupted positions.}
We derive the approximation by substituting the three values of the
clean noisy-vector coordinates into the exact ELBO.

For $m\ne y$,
\begin{equation}
    \bar{x}_{0,m}=1-\alpha_t,\qquad \bar{x}_{0,y}=V\alpha_t+1-\alpha_t,\qquad \bar{x}_{0,j}=1-\alpha_t\quad(j\ne y).
    \label{eq:app-cor1-clean-coordinates}
\end{equation}
Denote the pointwise integrand in Eq.~\eqref{eq:app-duo-weighted} by
$\mathrm{ELBO}$.  Substitution of
Eq.~\eqref{eq:app-cor1-clean-coordinates} and collection of the logarithms give
its model-dependent part
\begin{equation}
    \mathrm{ELBO}_{\theta} =-\alpha_t'\!\left[ \frac{1}{\alpha_t\bar{x}_{\theta,m}} +\frac{\log\bar{x}_{\theta,m}}{\alpha_t(1-\alpha_t)}
    -\frac{\log\bar{x}_{\theta,y}}{1-\alpha_t} -\frac{1}{V\alpha_t}\sum_{i=1}^{V}\log\bar{x}_{\theta,i} \right].
\end{equation}
With $R(z):=z^{-1}-1+\log z$,
\begin{equation}
    \frac{1}{\alpha_t\bar{x}_{\theta,m}} =\frac{1}{\alpha_t} -\frac{\log\bar{x}_{\theta,m}}{\alpha_t} +\frac{R(\bar{x}_{\theta,m})}{\alpha_t}.
\end{equation}
Removing the first, $\theta$-independent term yields
\begin{equation}
    -\alpha_t'\!\left[ \frac{\log\bar{x}_{\theta,m}-\log\bar{x}_{\theta,y}}{1-\alpha_t} -\frac{1}{V\alpha_t}\sum_{i=1}^{V}\log\bar{x}_{\theta,i}
    +\frac{R(\bar{x}_{\theta,m})}{\alpha_t} \right].
    \label{eq:app-cor1-with-remainder}
\end{equation}
The assumption $x_{\theta,m}=1/V+o(1/V)$ implies
\begin{equation}
    \bar{x}_{\theta,m}-1=\alpha_t(Vx_{\theta,m}-1)=o(1).
\end{equation}
Since $R(1+u)=u^2/2+O(u^3)$, the last term in
Eq.~\eqref{eq:app-cor1-with-remainder} is uniformly $o(1)$ on the stated
interior-time region.  The same remains true after averaging.

Define
\begin{equation}
    \mathrm{CE}_{\mathrm{LS}}^{1-\alpha_t}(\bar{\vx
    x}_{\theta},y)=-\left[\alpha_t\log\bar{x}_{\theta,y}+\frac{1-\alpha_t}{V}\sum_{i=1}^{V}\log\bar{x}_{\theta,i}\right].
\end{equation}
A direct rearrangement yields
\begin{equation}
    \frac{\log\bar{x}_{\theta,m}-\log\bar{x}_{\theta,y}}{1-\alpha_t}-\frac{1}{V\alpha_t}\sum_{i=1}^{V}\log\bar{x}_{\theta,i}=\frac{\mathrm{CE}_{\mathrm{LS}}^{1-\alpha_t}(\bar{\vx
    x}_{\theta},y)}{\alpha_t(1-\alpha_t)}+\frac{\log\bar{x}_{\theta,m}}{1-\alpha_t}.
\end{equation}

\paragraph{Already-clean positions.}
We show that $m=y$ contributes only $o(1)$ under the additional assumption.
Setting $m=y$ in Eq.~\eqref{eq:app-duo-weighted} and omitting
$\theta$-independent terms gives
\begin{equation}
    -\alpha_t'\!\left[ \frac{1}{\alpha_t\bar{x}_{\theta,y}} +\frac{1-\alpha_t}{V\alpha_t(V\alpha_t+1-\alpha_t)} \sum_{j\ne
    y}\log\frac{\bar{x}_{\theta,y}}{\bar{x}_{\theta,j}} \right].
\end{equation}
On the interior-time region,
$\varepsilon\leq\bar{x}_{\theta,j}\leq V$ and
$V\alpha_t+1-\alpha_t\geq V\varepsilon$.  Hence
\begin{equation}
    \frac{1}{\alpha_t\bar{x}_{\theta,y}} =O\!\left(\frac{1}{Vx_{\theta,y}}\right),\qquad \left| \frac{1-\alpha_t}{V\alpha_t(V\alpha_t+1-\alpha_t)}
    \sum_{j\ne y}\log\frac{\bar{x}_{\theta,y}}{\bar{x}_{\theta,j}} \right| =O\!\left(\frac{\log V}{V}\right).
\end{equation}
If $x_{\theta,y}=\omega(1/V)$, both terms are $o(1)$.  Boundedness of
$|\alpha_t'|$ completes the proof.

\subsubsection{Proof of Theorem 1}
\label{app:proof-cor2}

We first state the exact centered identity.  The proof then extracts the clean
coordinate of the learned reverse posterior, takes the small-step limit, and
ends with an explicit finite-step error bound.

\paragraph{Theorem 1 (Restated).}
Fix $\vx x_0=\vx e_y$, a corrupted state $\vx x_t=\vx e_m$ with $m\ne y$,
and a time satisfying $0<\alpha_t<1$.  For $\Delta t>0$, let
$s=t-\Delta t$ and assume $0<\alpha_t<\alpha_s<1$.  Under the Duo
$x_0$-parameterization, substituting the fixed model output
$\vx x_\theta(\vx x_t,t)$ into the exact reverse posterior gives the centered
identity
\begin{equation}
    \fitbox{0.98\textwidth}{$\displaystyle \KL\!\left(\delta(\vx x_0)\middle\|p_\theta(\vx x_s\mid\vx x_t)\right)
    +\log\frac{\alpha_s-\alpha_t}{V\alpha_s} =\log\bar{x}_{\theta,m} -\log\!\left(V\alpha_sx_{\theta,y}+1-\alpha_s\right).$}
    \tag{\number\numexpr\value{equation}+1\relax--\number\numexpr\value{equation}+2\relax}
\end{equation}
\addtocounter{equation}{2}
If such admissible reverse steps exist for arbitrarily small $\Delta t$, then,
as $\Delta t\to0$ along these steps,
\begin{equation}
    \KL\!\left(\delta(\vx x_0)\middle\|p_\theta(\vx x_s\mid\vx x_t)\right) +\log\frac{\alpha_s-\alpha_t}{V\alpha_s} \longrightarrow
    \log\bar{x}_{\theta,m}-\log\bar{x}_{\theta,y}.
\end{equation}

\paragraph{Proof.}
We compute the reverse probability of the single target token and then take its
small-step limit.  For any categorical distribution $q$,
\begin{equation}
    \KL(\delta_{\vx e_y}\|q)=-\log q(\vx e_y).
    \label{eq:app-thm1-dirac-kl}
\end{equation}

Substitute $\vx x_\theta(\vx x_t,t)$ for $\vx x_0$ in
Eq.~\eqref{eq:app-uniform-posterior-vector}.  At coordinate $y$, the two terms
containing $\vx x_t$ vanish because $m\ne y$.  Therefore
\begin{equation}
    \fitbox{0.98\textwidth}{$\displaystyle p_\theta(\vx x_s=\vx e_y\mid\vx x_t=\vx e_m)
    =\frac{(\alpha_s-\alpha_t)x_{\theta,y}+(1-\alpha_t/\alpha_s)(1-\alpha_s)/V}{\bar{x}_{\theta,m}} =\frac{\alpha_s-\alpha_t}{V\alpha_s}
    \frac{V\alpha_sx_{\theta,y}+1-\alpha_s}{\bar{x}_{\theta,m}}.$}
\end{equation}
Taking $-\log$ and using Eq.~\eqref{eq:app-thm1-dirac-kl} gives
\begin{equation}
    \KL\!\left(\delta(\vx x_0)\middle\|p_\theta(\vx x_s\mid\vx x_t)\right) +\log\frac{\alpha_s-\alpha_t}{V\alpha_s} =\log\bar{x}_{\theta,m}
    -\log\!\left(V\alpha_sx_{\theta,y}+1-\alpha_s\right).
\end{equation}

The added logarithm is independent of $\theta$ and removes the divergent part
of the small-step KL.  The model output remains fixed at the input
$(\vx x_t,t)$ while $s\to t$.  Continuity gives
$V\alpha_sx_{\theta,y}+1-\alpha_s\to\bar{x}_{\theta,y}$, proving the stated
limit.  Since $(\alpha_s-\alpha_t)/(V\alpha_s)\to0$, the raw KL diverges; only
its finite model-dependent part remains.  Weighting this part by
$-\alpha_t'/(1-\alpha_t)$ gives Eq.~\eqref{eq:lu-loss} on corrupted positions.

For completeness, we quantify how close a finite reverse step is to this
limit.  Define the nonsingular contrast
\begin{equation}
    \ell_{\mathrm{LU},\Delta}:=\log\bar{x}_{\theta,m}-\log\!\left(V\alpha_sx_{\theta,y}+1-\alpha_s\right).
\end{equation}
For $\alpha\in[\eta,1-\eta]$,
\begin{equation}
    \left|\frac{d}{d\alpha}\log(V\alpha x_{\theta,y}+1-\alpha)\right| =\frac{|Vx_{\theta,y}-1|}{V\alpha x_{\theta,y}+1-\alpha} \leq\frac{1}{\eta}.
\end{equation}
The inequality follows by considering $Vx_{\theta,y}\geq1$ and
$Vx_{\theta,y}\leq1$.  If $|\alpha_u'|\leq M_\eta$ and
$\alpha_u\in[\eta,1-\eta]$ for $u\in[s,t]$, the mean-value theorem gives
\begin{equation}
    \left| \ell_{\mathrm{LU},\Delta} -\left(\log\bar{x}_{\theta,m}-\log\bar{x}_{\theta,y}\right) \right| \leq\frac{|\alpha_s-\alpha_t|}{\eta}
    \leq\frac{M_\eta}{\eta}\Delta t.
    \label{eq:app-thm1-finite-delta-bound}
\end{equation}

Restricting to times for which these conditions hold on
$[t-\Delta t,t]$, define
\begin{equation}
    \Loss_{\mathrm{LU},\Delta}^{\ell}:=\E_{t,\vx x_t}\!\left[\I_{m\ne y}\frac{-\alpha_t'}{1-\alpha_t}\ell_{\mathrm{LU},\Delta}\right].
\end{equation}
Since
$p(m\ne y\mid y,t)=(1-\alpha_t)(1-1/V)$,
Eq.~\eqref{eq:app-thm1-finite-delta-bound} implies
\begin{equation}
    \fitbox{0.98\textwidth}{$\displaystyle \left|\Loss_{\mathrm{LU},\Delta}^{\ell}-\Loss_{\mathrm{LU}}^{\ell}\right|
    \leq\E_t\!\left[\frac{M_\eta}{1-\alpha_t}\frac{M_\eta}{\eta}\Delta t\;p(m\ne y\mid y,t)\right] =\frac{M_\eta^2}{\eta}\Delta
    t\left(1-\frac{1}{V}\right) \leq\frac{M_\eta^2}{\eta}\Delta t.$}
\end{equation}
For $\alpha_t=1-t$, this is $\Delta t/\eta$; thus
$\Delta t\leq\epsilon\eta$ gives error at most $\epsilon$.

\subsection{Per-Token Time Embeddings}

We show that per-token times preserve the forward marginal and derive their
sampling-time initialization.

\paragraph{Marginal preservation.}
We average over the per-token time and recover the original uniform-state
marginal.

For $\alpha_t=1-t$, draw $\tau^i\sim q_t$ independently and
$R^i\mid\tau^i\sim\mathrm{Bernoulli}(\tau^i)$.  If $R^i=0$, keep the clean
token; otherwise draw uniformly from $\mcV$.  Therefore
\begin{equation}
    \fitbox{0.98\textwidth}{$\displaystyle p(x_t^i=v\mid x_0^i=y,t) =\E_{\tau^i\sim q_t}\!\left[(1-\tau^i)\I_{v=y}+\tau^i\frac{1}{V}\right]
    =\left(1-\E_{q_t}[\tau^i]\right)\I_{v=y}+\frac{\E_{q_t}[\tau^i]}{V}.$}
\end{equation}
If $\E_{q_t}[\tau^i]=t$, then
\begin{equation}
    p(x_t^i=v\mid x_0^i=y,t)=(1-t)\I_{v=y}+\frac{t}{V}=\alpha_t\I_{v=y}+\frac{1-\alpha_t}{V},
\end{equation}
which is Eq.~\eqref{eq:app-uniform-marginal-scalar}.  Independence across
positions preserves the factorized sequence law.  For a non-linear schedule,
the same proof requires $\E_{q_t}[\tau^i]=1-\alpha_t$.

For $0<t<1$, the implementation uses
\begin{equation}
    q_t(\tau)=\mathrm{Beta}(ct,c(1-t)), \qquad c=2,
\end{equation}
\begin{equation}
    \E_{q_t}[\tau]=\frac{ct}{ct+c(1-t)}=t.
\end{equation}
At the endpoints, take $q_0=\delta_0$ and $q_1=\delta_1$; finite-precision
clipping introduces only the corresponding endpoint perturbation.  Note that
$x_t^i=x_0^i$ does not imply $R^i=0$, because uniform replacement redraws the
clean token with probability $1/V$.

\paragraph{Sampling-time initialization.}
Prompt positions are fixed conditions and receive $\tau=0$, while every
non-prompt position is randomized.  Under $\tau\sim\mathrm{Beta}(1,1)$ and
$\text{Token is random}\mid\tau\sim\mathrm{Bernoulli}(\tau)$, Bayes' rule gives
\begin{equation}
    \tau\mid \text{Token is random}\sim\mathrm{Beta}(2,1).
\end{equation}
Thus Algorithm~\ref{alg:ludi_sample} uses $\mathrm{Beta}(2,1)$ as initialization.

\paragraph{Use during denoising.}
We use the same interpretation throughout denoising.

Committed positions receive $\tau^i=0$ and act as clean conditioning context;
positive-time positions remain denoising targets.  This yields the
confidence-based easy-to-hard sampler used by LUDI.

\endgroup

\section{Algorithm Details}
\label{app:algorithm-details}

\subsection{Training and Sampling}

\begin{algorithm}[t]
\caption{Training of LUDI}
\label{alg:ludi_train}
\SetAlgoLined
\SetKwInOut{Input}{Input}
\SetKwInOut{Output}{Output}
\SetKwFunction{Model}{Model}

\Input{Clean data $\mathbf{x}_0$, noise schedule $\alpha_t$, model $\mathbf{x}_\theta$}
\Output{Trained model parameters $\theta$}

\For{each training step}{
    Sample global time $t \sim \mathcal{U}(0,1)$\;
    \For{each token position $i$}{
        \tcp{\textcolor{DarkGreen}{sample token-wise corruption time}}
        $\tau^i \sim \mathrm{Beta}(2t,\, 2(1-t))$\;
        Corrupt $\mathbf{x}_0^i$: with probability $\tau^i$, replace by a uniform token $\rightarrow \mathbf{x}_t^i$\;
    }
    \tcp{\textcolor{DarkGreen}{model predicts clean tokens from noisy inputs and per-token times}}
    $\hat{\mathbf{x}}_0 = \Model(\mathbf{x}_t,\; \boldsymbol{\tau})$\;
    \tcp{\textcolor{DarkGreen}{compute LU loss on corrupted positions (Eq.~\ref{eq:lu-loss})}}
    $\mathcal{L} = \frac{-\alpha_t'}{1-\alpha_t}\!\sum_{i: \mathbf{x}_t^i \neq \mathbf{x}_0^i} \bigl( \log \bar{x}_{\theta,m}^i - \log \bar{x}_{\theta,y}^i \bigr)$\;
    Update $\theta$ using $\nabla_\theta \mathcal{L}$\;
}
\end{algorithm}

\begin{algorithm}[t]

\caption{Sampling of LUDI}
\label{alg:ludi_sample}
\SetAlgoLined
\SetKwInOut{Input}{Input}
\SetKwInOut{Output}{Output}
\SetKwFunction{Model}{Model}

\Input{Prompt tokens, trained model $\mathbf{x}_\theta$, scheduler function $h$, confidence threshold $\gamma$}
\Output{Generated clean sequence $\mathbf{x}$}

Initialize $\mathbf{x}$: prompt tokens kept unchanged, all other tokens sampled uniformly from $\mathcal{V}$\;
Initialize $\boldsymbol{\tau}$: for non-prompt tokens $\tau^i \sim \mathrm{Beta}(2,1)$, for prompt tokens $\tau^i=0$\;

\While{$\max(\boldsymbol{\tau}) > 0$}{
    $\hat{\mathbf{x}}_0 = \Model(\mathbf{x},\; \boldsymbol{\tau})$\;
    \tcp{\textcolor{DarkGreen}{compute next per-token times via scheduler $h$}}
    $\boldsymbol{\tau}_{\text{next}} = h(\boldsymbol{\tau}, \hat{\mathbf{x}}_0)$\;
    \tcp{\textcolor{DarkGreen}{For confidence-based scheduler: high-confidence tokens are committed}}
    \tcp{\textcolor{DarkGreen}{$c^\ell := \max(\hat{\mathbf{x}}_0^\ell)$ if $\tau^\ell>0$ else 0}}
    \tcp{\textcolor{DarkGreen}{$\tau_{\text{next}}^\ell=0$ if $c^\ell>\gamma$ or $\ell=\arg\max_{i}c^i$; otherwise $\tau_{\text{next}}^\ell=\tau^\ell$}}
    update $\mathbf{x}$ from $\mathbf{x}_{\tau}$ to $\mathbf{x}_{\tau_{\text{next}}}$ using the reverse transition (Eq.~\ref{eq:reverse-posterior})\;
    $\boldsymbol{\tau} \leftarrow \boldsymbol{\tau}_{\text{next}}$\;
}
\Return $\mathbf{x}$\;
\end{algorithm}

The training procedure of LUDI is summarized in Algorithm~\ref{alg:ludi_train}.
It introduces per-token time embeddings $\boldsymbol{\tau}$ to provide token-level corruption hints,
and optimizes the less uniform (LU) loss (Eq.~\ref{eq:lu-loss}) that directly encourages each reverse
transition toward the clean token. The conditional sampling, detailed in Algorithm~\ref{alg:ludi_sample},
leverages a confidence-based scheduler $h$ to progressively commit high-confidence tokens as conditions,
enabling few-step generation for complex reasoning tasks.

\subsection{Details of AR to Block Diffusion}
\label{app:block-udlm-mask}
\paragraph{Label shift.}
AR pretraining predicts $x^\ell$ from a clean prefix $x^{<\ell}$, while block diffusion forces the model to predict $x^\ell$ at position $\ell$.
We resolve this mismatch by shifting the labels left by one position, while still allowing token $\ell-1$ to attend to token $\ell$.
This changes only the prediction position, preserving both the diffusion paradigm and the autoregressive prediction preference.

\paragraph{Complementary views.}
Following Fast dLLM v2~\citep{wu2025fast}, we train with two complementary mask patterns per sequence: positions kept clean in the first view are masked in the second, and vice versa.
In the UDLM setting, unmasked positions receive $\mathbf{x}_0$ and masked positions receive a random token.
Both views share a single forward pass, forcing every position to serve as both a clean target and a noisy input.
This doubling of supervision is valuable in our low-data regime.

\begin{figure}[t]
\centering
\begin{subfigure}[t]{0.27\textwidth}
    \centering
    \includegraphics[width=\linewidth]{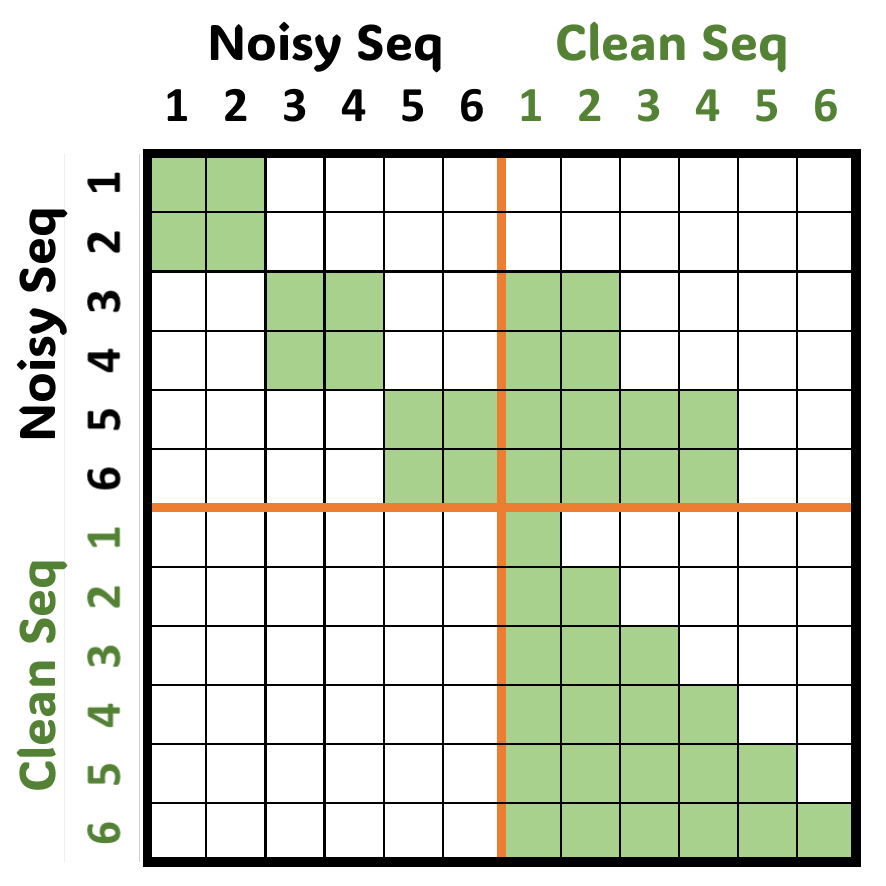}
    \caption{Training mask.}
    
\end{subfigure}
\hspace{25pt}
\begin{subfigure}[t]{0.27\textwidth}
    \centering
    \includegraphics[width=\linewidth]{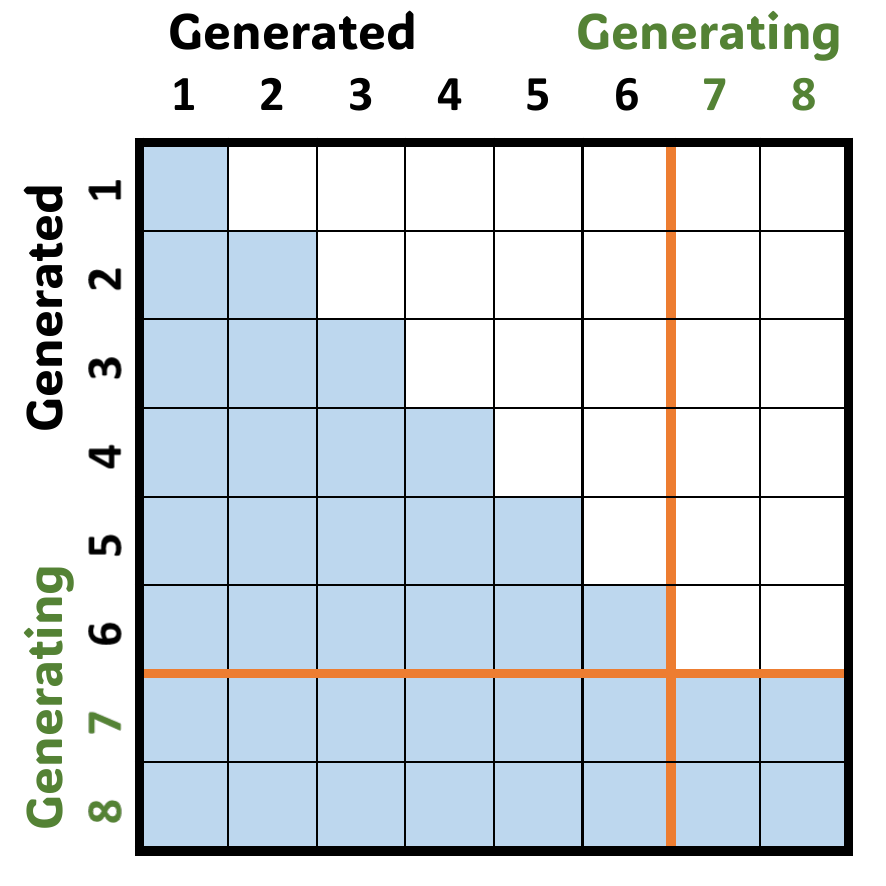}
    \caption{Inference mask.}
    
\end{subfigure}
\caption{Context-Causal attention masks for block-diffusion adaptation.
(a) Training mask: the prefix uses a lower-triangular causal mask; the active block attends bidirectionally within itself and causally to the prefix.
(b) Inference mask: generated blocks form a causal prefix; only the current block is denoised bidirectionally.}
\label{fig:context-causal-attention}
\end{figure}

\paragraph{Context-causal attention mask.}
NBDiff~\citep{tian2025nbdiff} proposed the context-causal attention mask.
The prefix observes a strictly causal mask, while the active denoising block enjoys full intra-block bidirectionality and causal access to the prefix.
This yields the attention mask pattern shown in Figure~\ref{fig:context-causal-attention}, where both noisy and clean sequences are used during training.
The clean sequence provides keys and values from causal attention for the subsequent noisy block.
This preserves the left-to-right inductive bias in already committed context, preventing instability from premature future exposure.
At inference the same pattern holds (Figure~\ref{fig:context-causal-attention}(b)): the prompt and past blocks form a frozen causal prefix, and only the current block is denoised bidirectionally, enabling KV-cache reuse and parallel token decoding.

\paragraph{AR loss.}
An auxiliary autoregressive loss is computed on the clean sequence.
This incurs no extra forward cost and serves dual purposes: it prevents catastrophic forgetting of the pretrained model, and it supplies a stable gradient signal that counters the noisier diffusion loss early in training.
The objective is $\mathcal{L} = \mathcal{L}_{\text{diff}} + \lambda \mathcal{L}_{\text{AR}}$ with a weight $\lambda$.

\section{Experiments}

\subsection{Experimental Setup for Small-Scale LUDI}
\label{app:small-settings}

\paragraph{170M models.}
We train four 170M UDLMs on OpenWebText~\citep{Gokaslan2019OpenWeb}: LUDI with
the LU loss, and baselines trained with Duo~\citep{sahoo2025duo},
SDDLM~\citep{zhu2025sddlm}, and SDDLM-v1~\citep{zhu2025sddlm} losses.  The
denoiser is a 12-block Transformer with hidden size 768, 12 attention heads, a
128-dimensional conditioning embedding, sequence length 1024, dropout 0.1, the
GPT-2 vocabulary of 50{,}257 tokens~\citep{Radford2019LanguageMA}, and bfloat16
precision. Each
run is trained for 500K iterations with batch size 512.  We use AdamW with learning rate $3\times10^{-4}$, betas $(0.9,0.999)$. Generative perplexity is computed with GPT-2
Large~\citep{Radford2019LanguageMA}, and sample diversity is measured by mean
token-level entropy in bits.  As in RADD~\citep{ou2025radd}, Gumbel-based
categorical sampling is performed in float64 precision.  For released
masked-diffusion baselines, we evaluate RADD~\citep{ou2025radd} and
MDLM~\citep{sahoo2024mdlm} checkpoints trained on the same dataset for 400K and
1M steps, respectively, using the same generation metrics.

\paragraph{1B models.}
We train four 1B-parameter UDLMs from scratch on FineWeb~\citep{penedo2024fineweb}
with the LLaMA tokenizer~\citep{touvron2023llamaopenefficientfoundation}.  The
denoiser is a 20-block Transformer with hidden size 2048, 16 attention heads, MLP
ratio 3.5, a 128-dimensional conditioning embedding, sequence length 2048,
dropout 0.0, a 32{,}000-token vocabulary and bfloat16 precision.  Following the SMDM implementation~\citep{nie2025scaling}, we mix
1\% variable-length FineWeb examples into training. Each run is trained for 500K iterations with batch
size 256. We
use AdamW with learning rate $2\times10^{-4}$, betas $(0.9,0.95)$, weight decay 0.1.  We
evaluate likelihood-based multiple-choice accuracy on PIQA~\citep{bisk2020piqa},
SIQA~\citep{sap-etal-2019-social}, ARC-Easy~\citep{clark2018think},
HellaSwag~\citep{zellers-etal-2019-hellaswag},
OpenBookQA~\citep{mihaylov-etal-2018-suit}, and
RACE~\citep{lai-etal-2017-race}.  Following SDDLM~\citep{zhu2025sddlm}, each
candidate answer is scored with the ELBO-based likelihood estimate induced by
Eq.~\eqref{eq:duo-loss}.  We report accuracy
(\texttt{acc}) for PIQA, SIQA, ARC-Easy, and RACE, and length-normalized
accuracy (\texttt{acc\_norm}) for HellaSwag and OpenBookQA.

\subsection{Experimental Setup for LUDI-7B}
\label{app:7b-settings}

\paragraph{Baselines.}
We compare LUDI-7B against autoregressive (AR) baselines and state-of-the-art diffusion language models at comparable scales.
AR baselines include Qwen2.5-7B-Instruct~\citep{qwen2.5} (the initialization checkpoint). Masked diffusion baselines include Dream~7B~\citep{ye2025dream} (adapted from Qwen2.5-7B), LLaDA~8B~\citep{nie2025llada}, LLaDA-1.5~8B~\citep{llada1.5}, LLaDA-MoE~7B~\citep{lladamoe} and Fast-dLLM v2~\citep{wu2025fast}.

\paragraph{Evaluation benchmarks.}
We evaluate on a diverse set of tasks spanning code generation, mathematical reasoning, instruction following, and knowledge-intensive question answering.
For code generation, we use HumanEval~\citep{humaneval} and HumanEval+~\citep{evalplus}, as well as MBPP~\citep{mbpp} and MBPP+~\cite{evalplus}.
Mathematical reasoning is assessed on GSM8K~\cite{gsm}. Instruction following is measured with IFEval~\cite{ifeval}. For knowledge-intensive tasks, we adopt MMLU~\cite{mmlu} and GPQA~\cite{rein2023gpqa}. All code benchmarks are evaluated using the EvalPlus framework~\cite{evalplus}.
For all tasks we report standard accuracy metrics following the evaluation protocol of Fast-dLLM v2~\cite{wu2025fast}.

\subsection{Mathematical Reasoning}
\label{app:math-analysis}

We supplement the mathematical-reasoning evaluation with results on MATH500~\cite{lightman2024let}, as reported in Table~\ref{tab:math-data}. Under the Dolci training setting, both UDLM and MDLM variants obtain a substantially lower score on MATH500. This pattern is shared across diffusion formulations and is therefore not specific to LUDI.

We attribute this shared behavior primarily to a distribution mismatch between the mathematical content in Dolci-Instruct-SFT and the evaluation benchmarks. The relevant sequence-length statistics are summarized below:

\begin{itemize}
    \item \textbf{Training data.} Mathematical examples in Dolci-Instruct-SFT are typically long, scenario-based questions with detailed explanations, with a median question length of 920 tokens. Mathematical examples in Nemotron-Post-Train v2, many of which originate from DeepSeek-R1, are more concise, with a median question length of 217 tokens.
    \item \textbf{Evaluation data.} Questions in GSM8K and MATH500 have median lengths of 257 and 143 tokens, respectively.
\end{itemize}

\begin{table}[t]
\centering
\small
\captionsetup{justification=centering}
\caption{Mathematical-reasoning accuracy with Dolci and Dolci-Nemo mixture.}
\label{tab:math-data}
\setlength{\tabcolsep}{5pt}
\renewcommand{\arraystretch}{1.1}
\begin{tabular}{@{}lcccc@{}}
    \toprule
    & \multicolumn{2}{c}{\textbf{GSM8K}} & \multicolumn{2}{c}{\textbf{MATH500}} \\
    \cmidrule(lr){2-3}\cmidrule(lr){4-5}
    \textbf{Method} & \textbf{Dolci} & \textbf{Dolci-Nemo} & \textbf{Dolci} & \textbf{Dolci-Nemo} \\
    \midrule
    SDDLM & 74.7 & 80.8 & 48.8 & 51.9 \\
    SDDLM-v1 & 73.5 & 80.8 & 48.1 & 51.9 \\
    Fast-dLLM v2 & 71.3 & \textbf{83.7} & \textbf{51.1} & 51.7 \\
    \rowcolor{ludigreen} LUDI & \textbf{75.2} & 82.1 & 49.0 & \textbf{52.6} \\
    \bottomrule
\end{tabular}
\end{table}

Nemotron-Post-Train v2~\cite{NemotronPostTrainingDatasetV2} therefore provides mathematical examples whose length and task style more closely match the evaluation distribution. To reduce the mismatch while keeping the training-set size comparable, we construct \textsc{Dolci--Nemo Mix}: we remove the Other, Tool Use, Safety, and Multilingual categories from Dolci-Instruct-SFT and replace them with mathematics, code, and STEM examples from Nemotron-Post-Train v2.

As shown in Table~\ref{tab:math-data}, introducing Nemotron-Post-Train v2 consistently improves mathematical reasoning across the compared objectives. LUDI improves from 75.2 to 82.1 on GSM8K and from 49.0 to 52.6 on MATH500, achieving the strongest MATH500 result among the compared methods. Other methods also improve on both benchmarks. The gains across both UDLM and MDLM variants support the distribution-mismatch analysis and show that LUDI benefits effectively from task-aligned mathematical data.

A characteristic failure mode in diffusion generation is repetitive looping, in which the model repeatedly emits the same span. The following GSM8K output illustrates this behavior:

\begin{tcolorbox}[
  enhanced,
  breakable,
  colback=green!2,
  colframe=green!55!black,
  boxrule=0.5pt,
  arc=1pt,
  left=4pt,
  right=4pt,
  top=4pt,
  bottom=4pt
]

\textbf{Question:} A food caterer was told to prepare gourmet hot dogs for 36 guests. While most people would only eat one hotdog, he prepared enough for half of the guests to be able to have two hotdogs. However, 40 guests showed up, and everyone wanted a second hotdog. How many guests did not get a second hotdog?

$\quad$

\textbf{Generated answer:} The caterer prepared enough hot dogs for half of the guests to have two hot dogs. This means that $36 / 2 = 18$ guests would have two hot dogs. The caterer prepared enough hot dogs for half of the guests to have two hot dogs. This means that $36 / 2 = 18$ guests would have two hot dogs. \ldots

$\quad$

\emph{The two sentences repeat approximately 55 times, producing approximately 1,634 words.}
\end{tcolorbox}

Among the incorrect MATH500 responses produced by models trained on Dolci, repetitive loops account for 21.5\% of LUDI's errors, compared with 27.8\% for the MDLM baseline Fast-dLLM v2. LUDI thus moderately reduces this diffusion-specific failure mode. Its uniform-state formulation allows generated tokens to be revised rather than irrevocably committed, which can help interrupt locally self-reinforcing spans.

\begin{table}[!t]
\centering
\small
\setlength{\tabcolsep}{3pt}
\renewcommand{\arraystretch}{1.15}
\begin{tabular}{@{}ccccccccc@{}}
\toprule
& \multicolumn{4}{c}{\textbf{LUDI}} & \multicolumn{3}{c}{\textbf{Autoregressive baseline}} & \\
\cmidrule(lr){2-5}\cmidrule(lr){6-8}
\shortstack{\textbf{Batch}\\\textbf{size}} &
\shortstack{\textbf{Latency}\\\textbf{(seconds)}} &
\shortstack{\textbf{Tokens}\\\textbf{per second}} &
\shortstack{\textbf{Tokens}\\\textbf{per step}} &
\shortstack{\textbf{Peak memory}\\\textbf{(GB)}} &
\shortstack{\textbf{Latency}\\\textbf{(seconds)}} &
\shortstack{\textbf{Tokens}\\\textbf{per second}} &
\shortstack{\textbf{Peak memory}\\\textbf{(GB)}} &
\shortstack{\textbf{Throughput}\\\textbf{speedup}} \\
\midrule
1  & 494.46 & 132.97 & 3.19  & 31.05 & 572.95 & 69.31  & 15.35 & \textbf{$1.92\times$} \\
4  & 371.99 & 229.20 & 5.76  & 31.46 & 227.70 & 176.32 & 15.56 & \textbf{$1.30\times$} \\
16 & 179.68 & 477.50 & 11.92 & 31.93 & 86.08  & 464.84 & 16.33 & \textbf{$1.03\times$} \\
32 & 106.97 & 572.60 & 14.20 & 32.95 & 63.43  & 629.86 & 17.39 & $0.91\times$ \\
\bottomrule
\end{tabular}
\caption{End-to-end efficiency comparison between LUDI (confidence threshold 0.7) and the autoregressive baseline Qwen2.5-7B-Instruct on GSM8K across batch sizes 1--32. Throughput speedup is computed as LUDI tokens per second divided by autoregressive tokens per second.}
\label{tab:wall-clock}
\end{table}

\subsection{End-to-End Inference Efficiency}
\label{app:wall-clock}

To provide a comprehensive efficiency profile, we conduct end-to-end efficiency evaluations on GSM8K for LUDI (confidence threshold 0.7) and the autoregressive baseline Qwen2.5-7B-Instruct at batch sizes 1, 4, 16, and 32. The results are summarized in Table~\ref{tab:wall-clock}.

From Table~\ref{tab:wall-clock}, we make three observations. (1) \textbf{Decoding efficiency.} LUDI achieves throughput speedups at batch sizes 1, 4, and 16 ($1.92\times$, $1.30\times$, and $1.03\times$, respectively), while the autoregressive baseline slightly leads at batch size 32 because of its more compact key-value cache. (2) \textbf{Memory.} Owing to parallel multi-token decoding, LUDI's peak memory is approximately 1.9--2.0$\times$ that of the autoregressive baseline. (3) \textbf{End-to-end latency.} LUDI exhibits higher latency because it tends to produce longer reasoning chains. This behavior reflects the training-data distribution: the median response length in Dolci's mathematics subset is 2,114 tokens, compared with 217 tokens for GSM8K.

In practice, the theoretical $3\times$ speedup is not fully realized end to end. Inference efficiency can be further improved through dLLM-specific infrastructure. These system-level techniques are orthogonal to our modeling contributions, and we leave their integration to future work.
\begin{figure}[t]
    \centering
    \begin{subfigure}[t]{0.49\textwidth}
        \centering
        \includegraphics[width=\linewidth]{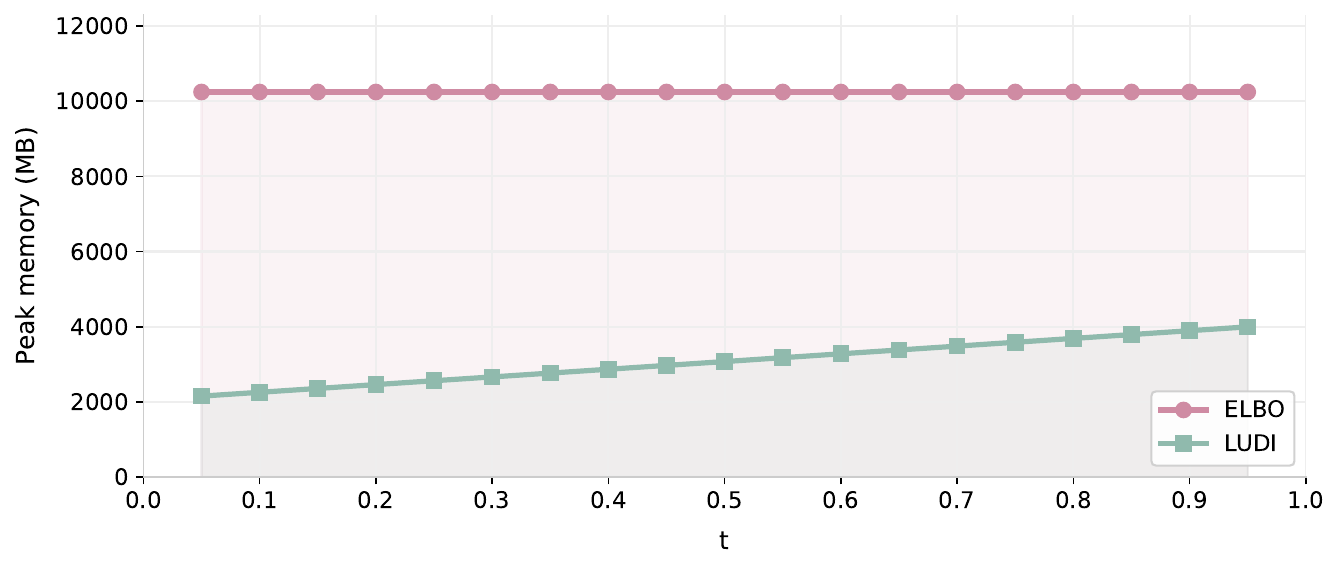}
        \caption{GPU Memory Usage.}
        
    \end{subfigure}
    \hfill
    \begin{subfigure}[t]{0.49\textwidth}
        \centering
        \includegraphics[width=\linewidth]{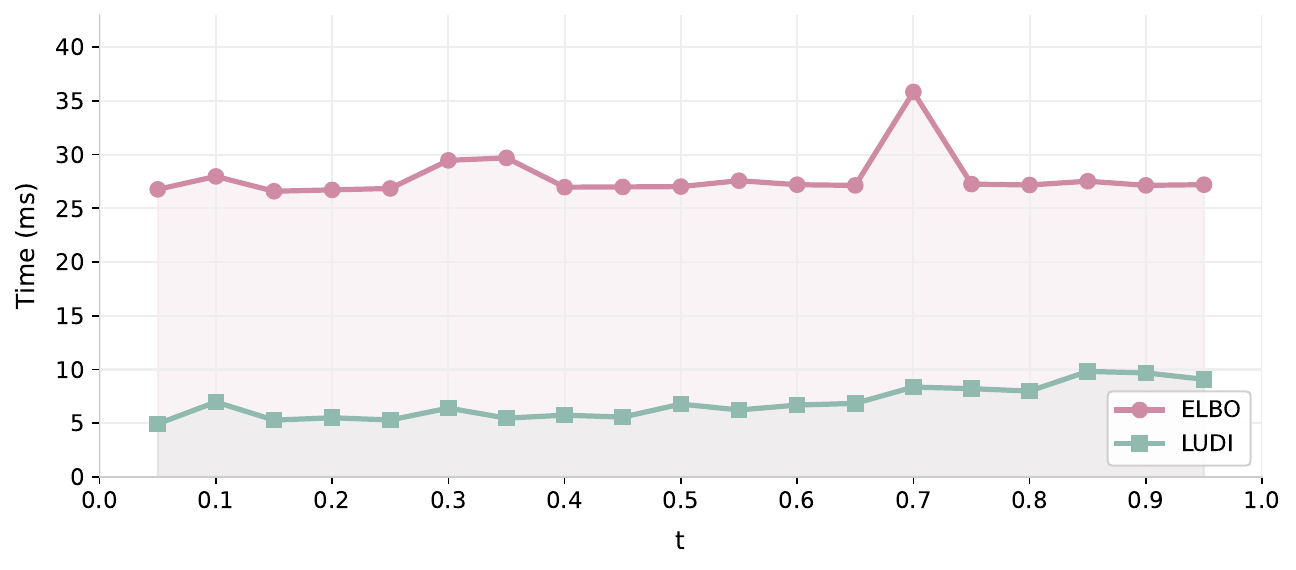}
        \caption{Forward and Backward Time.}
        
    \end{subfigure}
    \caption{Computational efficiency simulation in a long-sequence, large-vocabulary setting with batch size 10, sequence length 1024, and vocabulary size 50{,}000.}
    \label{fig:lu-loss-efficiency}
    \end{figure}
\begin{figure}[t]
    \centering
    \begin{subfigure}[t]{0.485\textwidth}
        \centering
        \includegraphics[width=\linewidth]{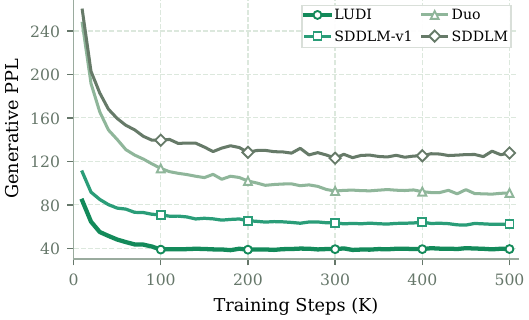}
        \caption{512 sampling steps.}
    \end{subfigure}\hfill
    \begin{subfigure}[t]{0.485\textwidth}
        \centering
        \includegraphics[width=\linewidth]{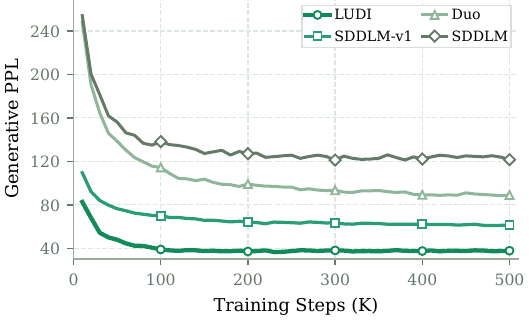}
        \caption{1024 sampling steps.}
    \end{subfigure}
    \caption{Generative perplexity over training checkpoints. Lower PPL is
    better. Lines connect all 10K-spaced checkpoints from 10K to 500K steps.}
    \label{fig:app-convergence}
\end{figure}
\subsection{Efficiency of the LU Loss}
\label{app:efficiency}

We explain why the LU loss in Eq.~\eqref{eq:lu-loss} admits a more efficient implementation than the ELBO loss in Eq.~\eqref{eq:duo-loss}. The gain comes from a simpler vocabulary-wide computation and from eliminating large intermediate tensors.

\paragraph{Fused implementation of the LU loss.}
Let $\mathbf{x}_\theta$ be the model prediction after softmax and recall that $\bar{x}_{\theta,i}=\alpha_tVx_{\theta,i}+(1-\alpha_t)$. The LU loss is
\begin{equation}
    \mathcal{L}_{\mathrm{LU}}^\ell=\frac{-\alpha_t'}{1-\alpha_t}\bigl(\log\bar{x}_{\theta,m}-\log\bar{x}_{\theta,y}\bigr).
\end{equation}
A direct implementation first materializes $\mathbf{x}_\theta$ and $\bar{\mathbf{x}}_\theta$ in $\mathbb{R}^{B\times L\times V}$ and then selects the entries at $m$ and $y$. Our Triton operator instead processes each token row as a stream. It computes the softmax normalization with an online log-sum-exp reduction, keeps the row statistics and the two selected logits in registers, and immediately evaluates $\log\bar{x}_{\theta,m}$ and $\log\bar{x}_{\theta,y}$. The full probability tensors therefore never need to be written to or read from global memory. The same compact row statistics are reused by the training operator, so normalization, index selection, and loss evaluation do not become separate framework operations.

\paragraph{Comparison with the ELBO loss.}
The ELBO loss additionally contains the vocabulary-wide term $\sum_{j=1}^{V}\log\bar{x}_{\theta,j}$. After obtaining the softmax normalization, it must still form $\bar{x}_{\theta,j}$, apply the logarithm, and reduce the result over every vocabulary entry. In a standard PyTorch implementation, these stages create several $B\times L\times V$ intermediates and launch multiple kernels. By comparison, LU performs the vocabulary scan needed for normalization and then applies the remaining nonlinear operations only at $m$ and $y$. Its fused implementation therefore uses fewer element-wise operations and reductions, launches fewer kernels, and transfers substantially less data between registers and global memory. These constant-factor savings are especially important for long sequences and large vocabularies. As shown in Figure~\ref{fig:lu-loss-efficiency}, the optimized LU operator achieves a $4.39\times$ speedup and a $3.45\times$ memory reduction relative to the PyTorch ELBO baseline.

\subsection{Convergence}

Figure~\ref{fig:app-convergence} compares the training trajectories of the four
UDLM objectives under the same evaluation protocol.  LUDI improves rapidly in
the early stage and remains in a low-perplexity regime as training proceeds.  At
500K steps, LUDI obtains the lowest generative perplexity under both 512-step
and 1024-step sampling, whereas Duo and SDDLM plateau at substantially higher
PPL.  These trajectories support the main-text observation that the LU loss
provides a cleaner and more sample-efficient training signal.

\subsection{Quality and Diversity}
\label{app:quality-diversity}

\begin{figure}[t]
    \centering
    \begin{subfigure}[t]{0.485\textwidth}
        \centering
        \includegraphics[width=\linewidth,trim=0 156 255.672 14,clip]{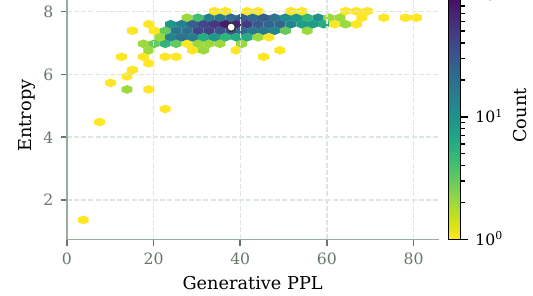}
        \caption{LUDI: Gen.PPL 37.9, entropy 7.50.}
    \end{subfigure}\hfill
    \begin{subfigure}[t]{0.485\textwidth}
        \centering
        \includegraphics[width=\linewidth,trim=255.672 156 0 14,clip]{figures/quality_diversity.pdf}
        \caption{SDDLM-v1: Gen.PPL 61.2, entropy 7.82.}
    \end{subfigure}
    \vspace{4pt}
    \begin{subfigure}[t]{0.485\textwidth}
        \centering
        \includegraphics[width=\linewidth,trim=0 0 255.672 170,clip]{figures/quality_diversity.pdf}
        \caption{Duo: Gen.PPL 89.0, entropy 7.98.}
    \end{subfigure}\hfill
    \begin{subfigure}[t]{0.485\textwidth}
        \centering
        \includegraphics[width=\linewidth,trim=255.672 0 0 170,clip]{figures/quality_diversity.pdf}
        \caption{SDDLM: Gen.PPL 123.1, entropy 8.10.}
    \end{subfigure}
    \caption{Density of per-sample generative PPL and sample entropy for
    the four losses at the 500K checkpoint, evaluated with 1024-step sampling
    and 1024 generated samples per loss. Each panel uses its own axis range and
    density scale for readability. Darker cells indicate more samples within the
    corresponding panel, and the white dot marks the method mean.}
    \label{fig:app-quality-diversity}
\end{figure}

Figure~\ref{fig:app-quality-diversity} compares the per-sample distributions of
the four losses at the 500K checkpoint, using 1024 denoising steps and 1024
samples for each loss. LUDI places a large fraction of its samples in the
low-Gen.PPL region, with a mean Gen.PPL of 37.9 and a mean entropy of 7.50.
Its samples are not concentrated in an extremely low-entropy region, suggesting
that the lower Gen.PPL is not accompanied by an obvious entropy collapse.

\end{document}